\documentclass[journal]{IEEEtran}

\usepackage{booktabs}
\usepackage{graphicx}

\usepackage{tabularx}
\usepackage{array}
\usepackage{booktabs}

\usepackage{float}

\usepackage{fancyhdr}
\usepackage{hyperref}

\ifCLASSINFOpdf
\else
\fi
\begin{document}
%
\title{Towards General Embodied Intelligence: Integrating Large Language Models, Knowledge Bases, and Reasoning Capabilities to Build the Next Generation of AI Agents}
%
%
%

\author{
Fujiang Yuan\textsuperscript{1}, 
Xia Huang\textsuperscript{1}, 
Lusheng Wang\textsuperscript{1}, 
Jun Ding\textsuperscript{1}, 
Zhen Tian\textsuperscript{2}, 
Yuxin Wang\textsuperscript{3}, 
Shaojie Gu\textsuperscript{4}, 
Yuki Funabora\textsuperscript{5}, 
Yanhong Peng\textsuperscript{1,5,*}, 
Zebing Mao\textsuperscript{6,*}

\thanks{
\textsuperscript{1} College of Mechanical Engineering, Chongqing University of Technology, Chongqing 400054, China\\
\textsuperscript{2} James Watt School of Engineering, University of Glasgow, G12 8QQ, UK\\
\textsuperscript{3} School of Energy and Power, Jiangsu University of Science and Technology, Zhenjiang 212100, China\\
\textsuperscript{4} Magnesium Research Center, Kumamoto University, Kumamoto 860-8555, Japan\\
\textsuperscript{5} Department of Information and Communication Engineering, Nagoya University, Nagoya 4648601, Japan\\
\textsuperscript{6} State Key Laboratory of Fluid Power and Mechatronic Systems, Zhejiang University, Hangzhou, 310027, China\\
\textsuperscript{*} Correspondence: yhpeng@nagoya-u.jp (Yanhong Peng), \\ mao.z.aa@zju.edu.cn (Zebing Mao).
}

}
\maketitle

\begin{abstract}
The convergence of large language models (LLMs), structured knowledge bases (KBs), and reasoning ability (RA) presents a promising trajectory toward general embodied intelligence (GEI). This paper reviews the evolution of LLM-centered intelligent systems, emphasizing their integration with knowledge representation, logical reasoning, and physical embodiment. We analyze LLM architectures, pre-training methods, and inference mechanisms, along with their interaction with external knowledge sources and structured reasoning frameworks. Furthermore, we examine embodied AI paradigms wherein agents learn and act in physical environments. A unified framework is proposed to illustrate the synergy among LLMs, KBs, RA, and embodiment, supporting perception, reasoning, and action. To advance toward GEI, we identify five key challenges: efficient LLM deployment, closed-loop knowledge integration, hybrid symbolic-neural reasoning, perception-action grounding, and continual learning. This survey provides a comprehensive roadmap for developing adaptive, multimodal agents capable of operating in complex, dynamic settings.
\end{abstract}


\IEEEpeerreviewmaketitle

\section{Introduction}
\IEEEPARstart{I}{n} recent years, large language models (LLMs) such as GPT-4, PaLM, Gemini, and Claude have emerged as powerful general-purpose learners, capable of performing a wide range of natural language tasks with remarkable fluency and coherence \cite{1,2}. Leveraging massive corpora and Transformer-based architectures, these models have demonstrated unprecedented performance in zero-shot and few-shot learning scenarios, ushering in a new paradigm in artificial intelligence. However, despite their impressive linguistic capabilities, LLMs remain fundamentally disembodied—they operate purely in the symbolic domain of language and lack grounded understanding of the physical or perceptual world \cite{3}.\par
At the same time, the field of embodied artificial intelligence (EAI)—which aims to endow agents (such as robots or virtual avatars) with the ability to perceive, act, and adapt in dynamic environments—has made significant strides \cite{4}. Modern embodied agents can perform real-time navigation, visual servoing, object manipulation, and task execution in both physical and simulated environments \cite{5}. Yet, these systems often rely on rigid, task-specific control pipelines and exhibit limited autonomy in complex decision-making, semantic comprehension, and long-horizon reasoning. This lack of high-level cognitive capability constrains their adaptability in open-ended, multi-step scenarios.\par

Bridging the cognitive strengths of LLMs with the sensorimotor embodiment of intelligent agents offers a promising path toward general-purpose interactive intelligence \cite{6}. By embedding LLMs into the control loops of embodied systems, researchers have begun to develop LLM-based agents that utilize natural language not only as an instruction interface but also as an internal medium for planning, memory management, tool use, and decision support. In this framework, LLMs act as “neural executors,” capable of interpreting complex commands, querying knowledge bases, reasoning over multimodal observations, and generating context-aware action plans \cite{7}.\par
Nevertheless, achieving such integration poses several fundamental challenges. A key issue is the symbol grounding problem: LLMs, trained solely on textual data, often struggle to translate abstract instructions into concrete sensorimotor actions in the real world. Additionally, these models are susceptible to hallucinations, suffer from limited memory, and lack explicit world models, which compromises their reliability in safety-critical or long-term reasoning tasks. While external structured knowledge sources—such as knowledge graphs, ontologies, and commonsense databases—can enhance the reasoning capabilities of LLMs, integrating them with large neural architectures remains an open technical bottleneck due to mismatches in representation, update mechanisms, and interfacing protocols.\par
Moreover, although the individual domains of LLMs, embodied intelligence, and neuro-symbolic reasoning have been actively explored, there remains a significant gap in systematizing their synergistic integration. The current literature lacks a unified perspective on how these fields can be cohesively combined to create autonomous, language-driven, knowledge-augmented embodied agents capable of multimodal interaction, long-term planning, and adaptive reasoning across diverse environments.\par
To address this gap, this survey provides a comprehensive review of the emerging interdisciplinary landscape at the intersection of large language models, embodied AI, and structured knowledge-based reasoning. Our primary contributions are as follows:\par

\subsection{Our Contributions}
In recent years, research on embodied intelligence has developed rapidly, and many review articles have emerged. As shown in Table 1. Yao et al. \cite{8} focused on the integration of multimodal models and robots, and sorted out embodied tasks such as navigation and question answering and the PPA paradigm; Dong et al. \cite{9} explored the application of LLMs in human-machine symbiotic manufacturing, emphasizing its improvement in interaction and collaboration capabilities; Tan et al. \cite{10} proposed the CEIM architecture to achieve the integration of data and semantics in customized manufacturing scenarios; Jeong et al. \cite{11} systematically analyzed the role of LLM/VLM in the five modules of robot control; Wang et al. \cite{12} evaluated the performance of GPT-4V in planning and reasoning tasks, and looked forward to its potential in scenarios such as agriculture and medical care.\par
Although the above studies have provided many insights into the integration of embodied intelligence and large language models, there are still some shortcomings: such as the lack of unified architecture design and module coordination mechanism, semantic misalignment between language and perception, resource constraints and reasoning bottlenecks in LLM deployment, and a general solution framework covering the entire process of perception-cognition-action has not yet been formed.\par

In response to the above challenges, we proposed a systematic review of "Towards General Embodied Intelligence" in this study. For the first time, we built a unified GEI technical architecture from the perspective of the coordination of language models, knowledge bases, reasoning mechanisms, and physical interaction systems, and clearly proposed five key research directions (lightweight deployment, knowledge closure, reasoning enhancement, perception alignment, and safety and control), and provided\par
\begin{table*}[t]
	\centering
	\caption{Comparative Summary of Related Embodied Intelligence Review Articles}
	\label{tab:embodied_review}
	
	\renewcommand{\arraystretch}{1.15}
	\setlength{\tabcolsep}{5pt}
	
	\begin{tabularx}{\textwidth}{
			>{\raggedright\arraybackslash}p{2.3cm}
			>{\centering\arraybackslash}p{1.2cm}
			>{\raggedright\arraybackslash}X
			>{\raggedright\arraybackslash}X
		}
		\toprule
		\textbf{Authors} &
		\textbf{Year} &
		\textbf{Core Contributions} &
		\textbf{Limitations} \\
		\midrule
		
		Yao et al. \cite{8} &
		2025 &
		Reviews multimodal embodied intelligence tasks and models; explores LLM/MLM integration with robots; proposes PPA paradigm and evaluates MLMs. &
		Classical AI lacks physical interaction; DRL is data-hungry and less robust; VLN training is costly and generalization is poor; LLMs lack causal reasoning; Sim2Real transfer is weak; evaluation standards are lacking. \\
		
		Dong et al. \cite{9} &
		2025 &
		Analyzes LLM-enabled Human-Robot Symbiotic Manufacturing (HRSM), including interaction, collaboration, and execution. &
		Privacy and data sensitivity risks; hallucination and bias in LLMs; absence of unified evaluation standards; integration difficulties in diverse environments; token length limits document processing. \\
		
		Tan et al. \cite{10} &
		2025 &
		Proposes CEIM to fuse multi-source data and LLM reasoning for semantic integration and decision-making in customized manufacturing. &
		Data fusion issues; low semantic alignment; real-time control accuracy is lacking; high computation cost; insufficient standardization of AI-device interfaces. \\
		
		Jeong et al. \cite{11} &
		2024 &
		Surveys five directions of LLM/VLM in robotics: reward design, control, planning, manipulation, and scene understanding. &
		Limited by embedded compute; weak sensor fusion; prompt bias and poor interpretability; lacks domain-specific knowledge; Sim2Real generalization is weak. \\
		
		Wang et al. \cite{12} &
		2025 &
		Discusses LLMs in robotics tasks and evaluates GPT-4V; explores future applications in agriculture and medicine. &
		Plan robustness is lacking; long prompts require expert input; fixed action sets reduce flexibility; API is closed-source; RGBD-based 3D reasoning is limited. \\
		
		\textbf{Our Survey} &
		\textbf{2025} &
		Systematically analyzed the synergy among LLMs, knowledge bases, reasoning capabilities, and embodied intelligence; proposed a unified framework and five future directions for GEI. &
		Limited empirical examples; insufficient discussion on multilingual settings and real-world deployment. \\
		
		\bottomrule
	\end{tabularx}
	
\end{table*}

a systematic technical roadmap. Compared with existing reviews, our research is more systematic, comprehensive, and forward-looking, providing theoretical support and technical direction for the development of the next generation of artificial intelligence agents.\par

\section{Related Technology Introduction}
In order to fully understand the construction path of general embodied intelligence, this section will systematically review and sort out the key technical components that constitute its foundation. The realization of general embodied intelligence depends on the deep collaboration between multiple heterogeneous subsystems, including powerful language understanding and generation modules, the representation and call mechanism of structured knowledge, the ability to reason and make decisions in complex environments, and the ability to perceive and act in the real or simulated world. Therefore, we will introduce it from four aspects: first, an overview of the current mainstream large language model and its ability boundaries in cognitive tasks; second, an analysis of the structural types and integration methods of the knowledge base (KB); third, an exploration of how multiple reasoning models provide logical consistency and long-range inference capabilities for intelligent agents; finally, a review of the current development status of embodied intelligence, especially its application potential in the perception and control closed loop. Through this systematic combing, we aim to provide a theoretical basis and technical background support for the collaborative integration and cross-modal modeling of each module in subsequent chapters.\par
The rest of this review is structured as follows: Section 2 systematically reviews the key technical foundations required to build General Embodied Intelligence (GEI), including four core components: large language models (LLMs), knowledge bases (KBs), reasoning capabilities (RA), and embodied intelligence (EI), and analyzes them from the perspectives of technical architecture, development history, and integration methods. Section 3 deeply explores the fusion mechanism of the above modules and constructs an overall architecture of an intelligent system with LLM as the core, integrating knowledge representation, structured reasoning, and perception control. Section 4 discusses the challenges and frontiers of embodied intelligence, including five key research directions: lightweight deployment, knowledge closure, reasoning enhancement, perception alignment, and security control. Section 5 summarizes this survey.\par

\subsection{Large Language Model}

With the rapid evolution of artificial intelligence technology, Large Language Models have gradually become one of the most representative foundations of general artificial intelligence. As the core module of "language understanding and generation" in embodied intelligent systems, LLMs not only carry the deep modeling capabilities of natural language, but also play a key role in cross-modal fusion, knowledge expression, task planning, etc. This section will systematically explain the large language model, first reviewing its basic concepts and core features, then reviewing its technical development context, analyzing its key technical architecture and mechanisms, and further introducing its construction and execution process, laying the foundation for understanding its supporting role in general embodied intelligent systems.\par

\subsubsection{Overview of Large Language Model}
The study of language modeling originated in the 1990s, initially using statistical learning methods to predict the next word through previous words. However, this method has certain limitations when dealing with complex rules and contexts in language. Subsequent research has continued to promote the development of language models. In 2003, Bengio, a pioneer of deep learning, first introduced the concept of deep learning into language models in his classic paper "A Neural Probabilistic Language Model" and used neural network models for language modeling \cite{13}. This innovation provides computers with more powerful "cognitive" capabilities, allowing the model to more effectively capture the deep relationships in language.\par
Although this progress has significantly improved the performance of language models, there is still room for further improvement. Around 2018, the neural network model of the Transformer architecture was introduced into the language modeling task and made breakthrough progress \cite{14}. With the help of massive text data for training, the Transformer model can deeply understand language rules and patterns from a large-scale corpus, just like a computer can "read" the entire Internet \cite{15}.\par
This method has performed well in many natural language processing tasks and significantly improved the machine’s ability to understand and generate language. As an important achievement in this field, the Large Language Model has had a profound impact on many fields such as natural language processing, information retrieval, and computer vision \cite{16,17}.\par
In the field of natural language processing, LLM has significantly improved the performance of computers in tasks such as text generation, question answering and machine translation ; in the field of information retrieval, the application of LLM has optimized search engines and greatly improved the efficiency of information search \cite{18}; in addition, researchers are also exploring the combination of LLM and computer vision to enhance multimodal interaction \cite{19}.\par
Most importantly, the rise of LLM has given people new thinking about the possibility of general artificial intelligence (AGI). AGI refers to artificial intelligence with human-like thinking and learning capabilities, which can perform complex cognitive tasks and self-improve like humans. LLMs have several notable features that have attracted widespread attention in natural language processing and related fields \cite{20,21}. LLMs have several notable features that have attracted widespread attention in natural language processing and related fields. The following are the main features of LLMs:\par
\noindent\textbullet\ Large scale: LLMs typically contain billions to hundreds of billions of parameters, which greatly enhances their ability to capture complex language knowledge and grammatical structures. The large scale of the model enables deep understanding of language nuances and complex patterns.\par
\noindent\textbullet\ Context-awareness: LLMs demonstrate strong context-awareness and can accurately generate and parse text based on previous information. This feature makes them outstanding in tasks such as dialogue systems, article generation, and contextual reasoning.\par
\noindent\textbullet\ Multimodal capabilities: Some LLMs have been extended to multimodal data processing, including text, images, and speech. This allows the model to understand and generate cross-media content and support a variety of application scenarios.\par
\begin{figure*}[t]
	\centering
	\includegraphics[width=0.96\linewidth]{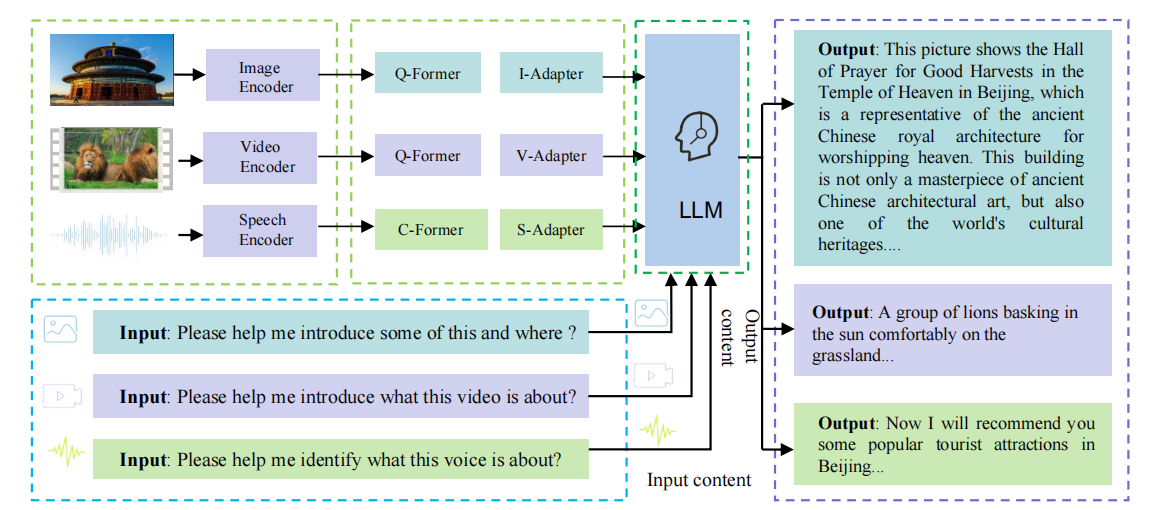}
	\caption{Schematic diagram of a unified multimodal large language model architecture.}
	\label{fig:figure1}
\end{figure*}
\noindent\textbullet\ Emergent capabilities: LLMs exhibit significant emergent capabilities, that is, as the model size increases, the model shows unprecedented performance improvements when handling more complex tasks. Emergent capabilities give LLMs a unique advantage in solving difficult problems.\par
Large language models are a technology with powerful language processing capabilities that have shown great application potential in many fields. It provides strong support for natural language understanding and generation tasks, but also triggers deep thinking about its ethical and risk issues. Therefore, LLM is not only an important research direction in the field of computer science and artificial intelligence, but also a key technology for future development. Figure 1 is a schematic diagram of the unified multimodal large language model architecture. This figure illustrates the processing flow of a multimodal LLM capable of handling diverse input modalities (image, video, speech) and generating natural language outputs. The architecture employs modality-specific encoders (Image Encoder, Video Encoder, Speech Encoder) to extract features from raw input data. These features are then aligned and integrated into the language model’s latent space using specialized adapter modules (I-Adapter for image features with Q-Former, V-Adapter for video features with Q-Former, S-Adapter for speech features with C-Former). The integrated multimodal information is processed by the core LLM to generate contextually relevant textual responses (Output). This unified architecture highlights the model’s ability to interpret heterogeneous inputs and produce coherent, task-oriented natural language responses, showcasing key advancements in multimodal understanding and generation.\par

\subsubsection{The Development of Large Language Models}

In the early days of computer science, researchers began to explore how to make machines understand and generate human language. Early natural language processing (NLP) methods mainly relied on rule-based systems and keyword matching, but this method had obvious limitations in dealing with complex language structures and understanding context. With the improvement of computing power and the advancement of algorithms, researchers gradually turned to statistical language models, such as n-gram models, which predict the next word by analyzing the probability of words in a large corpus. Although this improves keyword matching, it is still limited by data sparsity and contextual constraints \cite{22}.\par
With the rise of deep learning, NLP has entered a new era. Deep neural networks enable machines to learn more complex language patterns, get rid of the reliance on manual rules, and open up learning-based language modeling methods \cite{23}. In particular, the success of AlexNet in image recognition in 2012 demonstrated the potential of deep learning in various fields. Researchers subsequently extended it to NLP tasks and introduced models such as convolutional neural networks (CNNs) and recurrent neural networks (RNNs) \cite{24}, which can capture long-distance dependencies and simulate the temporal characteristics of language. However, RNN still faces the challenges of gradient vanishing and long-term dependency modeling \cite{25}. LSTM (Long Short-Term Memory Network) effectively alleviates these problems by introducing a gating mechanism, significantly improving the ability to process long sequences \cite{26}. Nevertheless, the sequential calculation of LSTM limits the efficiency of parallelization. The Transformer model proposed in 2017 completely changed this situation. Transformer abandons the recursive structure and adopts the self-attention mechanism, which significantly improves the ability to process long sequences and achieves efficient parallel computing. The self-attention mechanism enables the model to fully consider the entire input sequence when processing each word and capture long-distance dependencies, thereby performing well in complex tasks . Subsequently, the emergence of pre-trained language models such as BERT and GPT marked an important breakthrough in NLP. BERT uses a masked language model (MLM) for bidirectional context learning, while GPT uses an autoregressive approach for language modeling. These pre-trained models are trained on large-scale texts, significantly improving the machine’s ability to understand and generate language, and have achieved unprecedented results in multiple NLP tasks \cite{27}. Figure 2 shows the evolution and classification of Large Language Models.\par
This figure 2 illustrates the developmental trajectory and key characteristics of prominent Large Language Models over the period from 2018 to 2024. The models are categorized along three primary dimensions: (1) Release Timeline : Spanning from foundational models like BERT and GPT-1/2 to recent iterations such as GPT-4 and Chat GLM-4. (2) Model Availability : Distinguished as Open Source (e.g., BERT, BLOOM, LLaMA) or Closed Source (e.g., GPT-3, GPT-4, InstructGPT, ERNIE 3.0). (3) Model Architecture \& Primary Suitability : Models are grouped into three architectural paradigms: Encoder-only (e.g., BERT, ALBERT, ROBERTa, DistillBERT), optimal for Natural Language Understanding (NLU) tasks (e.g., text classification, sequence labeling, information extraction); Encoder-Decoder (e.g., BART, T5, mT5), well-suited for Natural Language Generation (NLG) tasks (e.g., text generation, machine translation, dialogue generation); and Decoder-only (e.g., GPT series, LLaMA, BLOOM, ChatGLM), demonstrating strong performance across a broad spectrum of Natural Language Processing (NLP) tasks, particularly text generation, machine translation, intelligent Q \& A, and increasingly sequence labeling. The progression highlights the trend towards decoder-only architectures dominating recent state-of-the-art models, alongside the coexistence of open-source and closed-source development models.\par

\subsubsection{Large Language Model Technical Architecture}

The Large Language Model is an important breakthrough in the field of Natural Language Processing. Its architecture and core technology are based on deep learning neural networks, especially the Transformer architecture. Since its introduction in 2017, the Transformer model has quickly become the core of language modeling with its innovative self-attention mechanism and efficient parallel computing capabilities \cite{28}. Unlike traditional recurrent neural networks (RNNs) and long short-term memory networks (LSTMs), Transformers can handle long-distance dependencies and avoid the gradient vanishing problem, greatly\par
\begin{figure*}[t]
	\centering
	\includegraphics[width=0.85\linewidth]{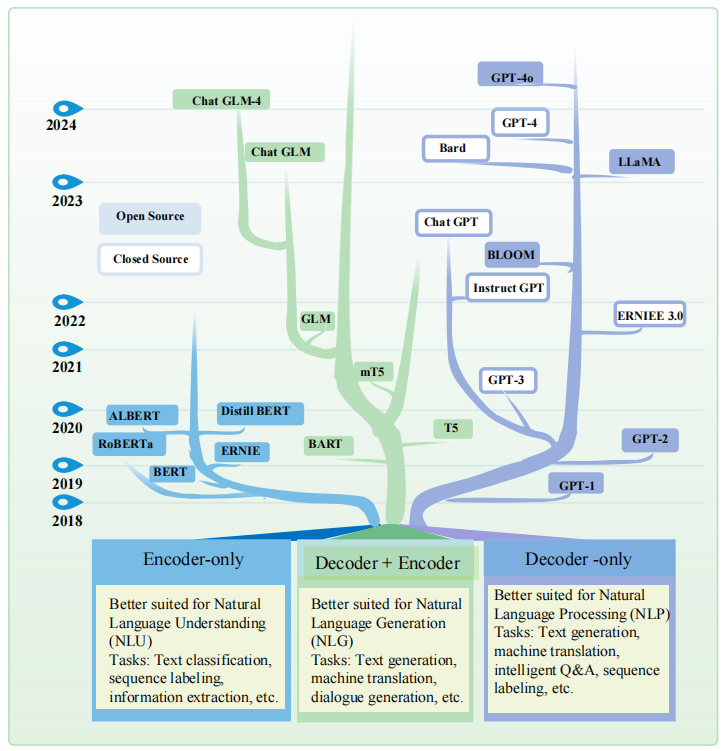}
	\caption{Evolution and Taxonomy of Large Language Models.}
	\label{fig:figure2}
\end{figure*}
improving the efficiency and accuracy of the model in long text processing. Its self-attention mechanism enables each word to dynamically adjust its weight based on the information of other words in the entire input sequence, thereby capturing complex contextual relationships \cite{29}.\par

This figure 3 illustrates the core structure of the Transformer neural network, foundational to modern large language models. Input tokens undergo Input Embedding and Positional Encoding to incorporate semantic and sequential information; an optional Additional Encoding layer may augment this representation. The Encoder stack (left) processes inputs through identical layers, each containing a Multi-Head Attention mechanism (capturing contextual dependencies), an Add \& Norm operation (residual connection with layer normalization), and a position-wise Feed Forward network followed by another Add \& Norm. For sequence generation tasks, the Decoder stack (right) utilizes Output Embedding (with positional encoding) of previous tokens. Each decoder layer comprises: Masked Multi-Head Attention (restricting attention to prior positions for autoregressive generation) with Add \& Norm; a Multi-Head Attention layer over encoder outputs (aligning source-target information) with Add \& Norm; a Feed Forward network with final Add \& Norm. The decoder output is projected via a Linear layer and normalized by Softmax to predict the next token probability distribution. Key innovations include
attention-based parallelism and stabilized training through residual connections (Add) and layer normalization (Norm).\par
The success of the Transformer architecture provides the basis for the rise of large language models. Large language models usually rely on pre-training and fine-tuning strategies. First, the model is pre-trained on a large corpus through unsupervised learning to learn the general representation of language. BERT uses the masked language model (MLM) method to learn bidirectional context by predicting masked words, while GPT uses an autoregressive model to predict the next word, thereby building a powerful language generation capability. The goal of the pre-training stage is to enable the model to gain a deep understanding of language structure, grammar, semantics, etc., through a large amount of data, while fine-tuning is to further adjust the model according to specific tasks so that it can efficiently cope with different NLP tasks, such as text classification, sentiment analysis, machine translation, etc. \cite{30}.\par

In addition, the multi-head attention mechanism of the large language model further enhances its expressive power, enabling the model to simultaneously learn different features in multiple subspaces, thereby improving the modeling ability of language patterns \cite{31}. It is also equipped with positional encoding, which provides the model with the position of words in the sequence, allowing Transformer to process sequence data without recursive structure \cite{32}. Through these technologies, the large language model can efficiently capture long-distance dependencies, process large-scale data, and perform parallel computing.\par
The powerful capabilities of the large language model also benefit from its huge parameter scale and large-scale training data. For example, GPT-3 has more than 175 billion parameters and can learn the deep structure and nuances of language through massive amounts of data \cite{33}. As computing power increases, the size of these models increases, and the demand for data and computing resources increases, but their performance becomes stronger and stronger.\par
In addition, in recent years, the Transformer architecture has expanded beyond text processing to multimodal learning, combining multiple data types such as text, images, and speech, further broadening its application scenarios. For example, models such as CLIP and DALL·E combine vision and language to enable large language models to not only understand text information, but also generate image-related descriptions and images \cite{34,35}.\par
Although large language models have shown great potential, they also face many challenges, including high computing resource consumption, ethical issues caused by data bias, and model interpretability. The training and reasoning processes require extremely large computing resources, which limits the popularity and application of the models \cite{36}. At the same time, the model may learn biases in the data during training, resulting in the generation of harmful or inaccurate content \cite{37}. In addition, due to the complexity of the model, the reasoning process is less transparent, which makes it difficult to effectively explain and understand the decision-making process of the model \cite{38,39}. Therefore, while further developing large language models, how to optimize computational efficiency, solve ethical issues, and improve model interpretability will be important research directions in the future.\par

\begin{figure*}[t]
	\centering
	\includegraphics[width=0.80\linewidth]{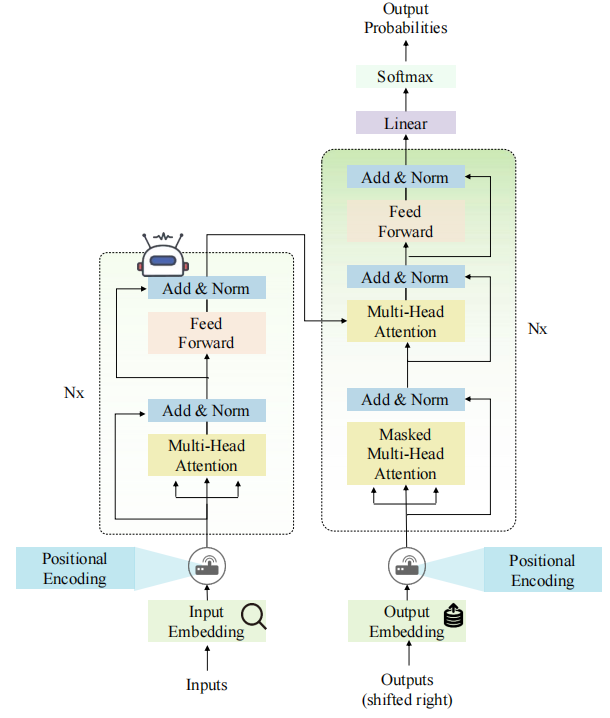}
	\caption{Schematic architecture of the Transformer model.}
	\label{fig:figure3}
\end{figure*}
\subsubsection{Large language model construction process and execution process}
The construction of a large language model follows a multi-stage process that integrates data engineering, algorithm design, infrastructure optimization, and iterative evaluation. This figure 4 outlines the standardized workflow adopted in developing modern LLMs, which typically includes five key phases: data collection and preprocessing, model architecture design, pre-training, fine-tuning, and evaluation.\par
1. Data Collection and Preprocessing The foundation of any LLM lies in the availability of large-scale, high-quality textual corpora. These datasets are curated from diverse sources such as Common Crawl, Wikipedia, academic literature, web forums, and domain-specific documents. Prior to training, raw text undergoes a series of preprocessing steps, including deduplication, language filtering, sentence segmentation, tokenization, and noise removal. For multilingual or domain-specific models, careful dataset balancing and ethical filtering are also performed to ensure representativeness and reduce bias.\par
2. Model Architecture Design Most modern LLMs are built on the Transformer architecture, which supports various configurations including encoder-only (e.g., BERT), decoder-only (e.g., GPT), and encoder-decoder (e.g., T5, mT5). The design process involves selecting model depth, width, number of attention heads, vocabulary size, and embedding strategies. Some models incorporate specialized components such as position encodings, adapter layers, or mixture-of-expert modules to support scalability or multimodal fusion.\par
3. Pre-training In this stage, the model is trained on massive unlabeled text using self-supervised learning objectives. Typical tasks include masked language modeling (MLM),\par
causal language modeling (CLM), or span prediction. Pre-training enables the model to learn general-purpose linguistic representations and semantic associations. Training requires large-scale distributed computing resources, often using hundreds or thousands of GPUs or TPUs, and may last for weeks. During this phase, strategies like gradient accumulation, mixed-precision training, and checkpoint sharding are applied to optimize memory and speed.\par
4. Fine-tuning and Alignment To adapt the general model to specific downstream tasks, fine-tuning is applied using labeled datasets. This may involve supervised learning (e.g., classification, summarization), reinforcement learning from human feedback (RLHF), or instruction tuning with curated prompts. For safety, controllability, and factuality, additional alignment steps may be included, such as reward modeling or prompt engineering. In multimodal settings, fine-tuning extends to vision-language or speech-text alignment.\par

\begin{figure}[t]
	\centering
	\includegraphics[width=0.96\linewidth]{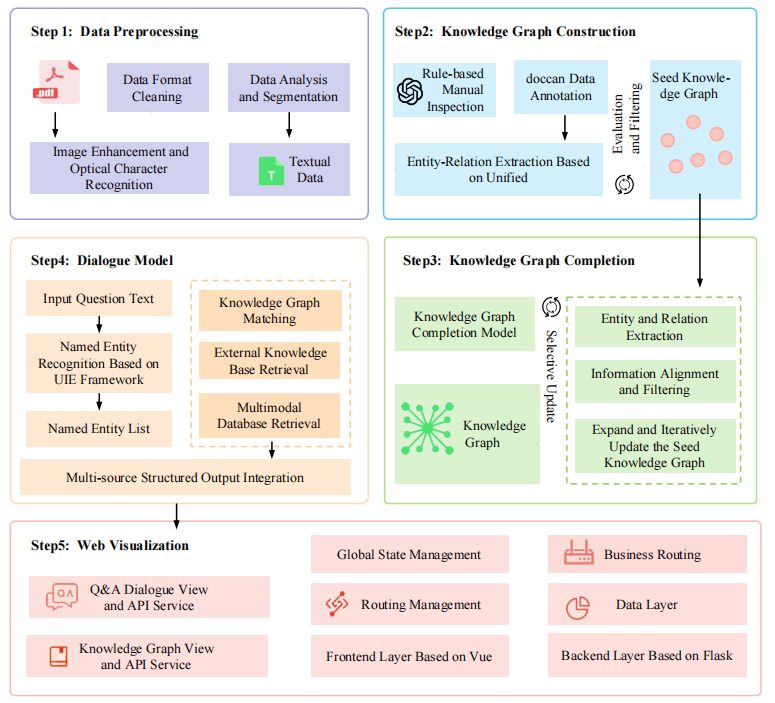}
	\caption{Construction Process of Large Language Models.}
	\label{fig:figure4}
\end{figure}
5. Evaluation and Iteration The trained LLM is evaluated using a suite of benchmarks covering language understanding (e.g., GLUE, SuperGLUE), generation (e.g., HELM, MMLU), reasoning (e.g., BIG-Bench), and task-specific metrics (e.g., BLEU, ROUGE, Exact Match). Robustness, fairness, and hallucination detection are also assessed. Based on evaluation results, iterative refinement is often performed by adjusting data, architecture, or learning objectives. Successful deployment further requires compression, quantization, and latency control for practical use in real-world environments.\par

While the construction pipeline defines the foundational components and training stages of LLMs, their practical value is ultimately realized through deployment in real-world applications. Once trained, these models are integrated into intelligent systems that leverage their capabilities for a wide range of downstream tasks. The following example illustrates how a fully constructed LLM can be embedded within an AI-powered document processing and question-answering (QA) framework to enable automated knowledge extraction and user interaction.\par
\begin{figure}[t]
	\centering
	\includegraphics[width=0.96\linewidth]{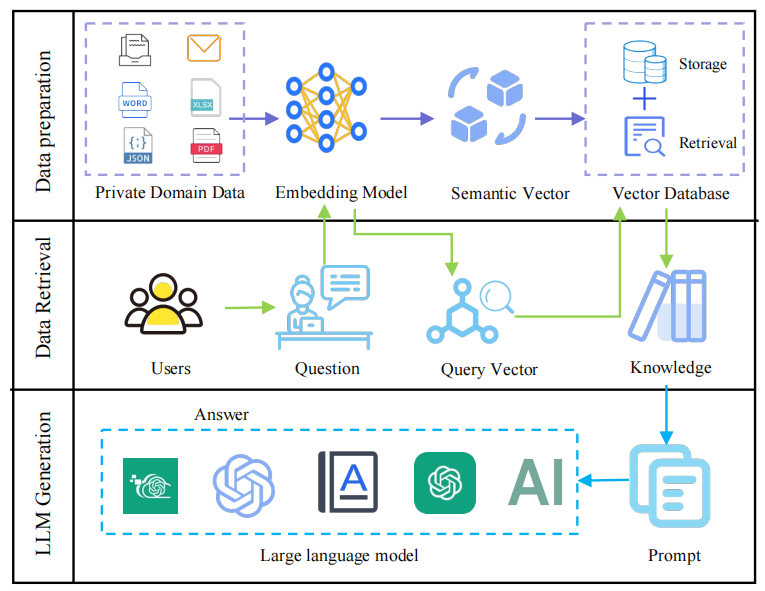}
	\caption{Description of the execution process of the intelligent document processing and knowledge question answering system.}
	\label{fig:figure5}
\end{figure}

Figure 5 show an end-to-end execution workflow example of an AI-driven document processing and question answering (QA) system. The process first ingests heterogeneous data formats, including Word, Excel, PDF, JSON, and email files. The system then parses these documents using deep learning models, extracts structured semantic information, and stores and indexes it into a centralized knowledge base. Upon receiving a user query, the system performs intent recognition and entity resolution to initiate relevant knowledge retrieval. The system leverages AI modules such as GPT-4, translation engines, and reasoning components to enhance semantic understanding and generate accurate responses. The final results are presented in structured or natural language form, returned to the user, and can be further archived for use by downstream applications. This workflow demonstrates a closed-loop process of "document-model-knowledge-interaction-generation", enabling scalable intelligent information services based on multimodal data and LLM capabilities.\par

\subsubsection{Current status of research on large language models}
The recent literature on Large Language Models highlights their growing role in a variety of applications, as well as the ongoing efforts to optimize their performance and scalability. In this context, Cummins et al. \cite{40} introduce the Meta Large Language Model Compiler, focusing on the development of a scalable and cost-effective foundation for compiler optimization tasks. Their work demonstrates the effectiveness of fine-tuned models in optimizing code size and enhancing disassembly processes, thereby showcasing LLMs’ potential in technical domains such as assembly code analysis. In the field of knowledge-based applications, Hu et al. \cite{41} explore prompting strategies that incorporate contextual information and pre-answers to improve LLM performance in Visual Question Answering (VQA), underscoring the critical role of context in enhancing model responses. Similarly, Hu et al. \cite{42}present LLM-TIKG, a framework for constructing Threat Intelligence Knowledge Graphs, further illustrating the capacity of LLMs to facilitate complex knowledge graph construction in cybersecurity contexts. Scalability and infrastructure requirements for training large models are addressed by Qian et al. \cite{43}, who describe Alibaba’s HPN network, specifically designed for large-scale LLM training in data centers. This infrastructure meets the growing computational demands of LLM development. Sun et al. \cite{44}contribute to the understanding of model scaling by introducing UniCoder, which employs an intermediate representation known as universal code (UniCode) to enhance the handling of downstream NLP tasks. Their work demonstrates how intermediate reasoning steps can significantly improve LLM performance. The application of LLMs in specialized scientific domains is exemplified by Zhang et al. \cite{45}, who develop BB-GeoGPT, a framework tailored for Geographic Information Science. This work illustrates the adaptability of LLMs to domain-specific knowledge and underscores their utility in specialized fields. In the realm of multimodal understanding, Wang et al. \cite{46} propose VideoAgent, an agent-based system that leverages LLMs to interpret long-form video content by integrating vision-language foundation models. This approach addresses the challenges of multimodal reasoning over extended sequences, enhancing LLM capabilities in video analysis. Further advancing multimodal capabilities, Xia et al. \cite{47} introduce LLMGA, a generation assistant that integrates LLMs with image generation and editing tools. Unlike traditional methods that generate fixed embeddings, LLMGA enables precise control over image synthesis through detailed language prompts, exemplifying the integration of LLMs with visual modalities. Methodological innovations in model alignment and evaluation are presented by Zhou et al. \cite{48}, who propose weak-to-strong search, a technique that aligns large models with smaller, fine-tuned counterparts through a greedy search process. This method offers a compute-efficient alternative to direct tuning and improves model generalization during testing. Additionally, Land et al. \cite{49} investigate the issue of ’glitch tokens’—tokens that are under-trained or absent during training—and propose a comprehensive detection method to enhance tokenizer robustness. Together, these studies underscore the versatility of LLMs across technical, scientific, and multimodal domains, while addressing critical challenges related to scalability, training infrastructure, model alignment, and tokenization.\par

\subsection{Knowledge base}

The knowledge base of large language models (LLMs) is not a monolithic component, but rather a dynamic system composed of multiple interrelated modules. As illustrated in Figure 6, the construction and utilization of an LLM’s knowledge base involves a comprehensive\par
\begin{table*}[t]
	\centering
	\caption{Comparison of Research on Large Language Models (LLMs)}
	\label{tab:llm_comparison}
	
	\small
	\renewcommand{\arraystretch}{1.08}
	\setlength{\tabcolsep}{4pt}
	
	\begin{tabularx}{\textwidth}{
			>{\raggedright\arraybackslash}p{2.2cm}
			>{\raggedright\arraybackslash}X
			>{\raggedright\arraybackslash}X
			>{\raggedright\arraybackslash}p{3.0cm}
		}
		\toprule
		
		\textbf{Author(s)} &
		\textbf{Core Improvement} &
		\textbf{Advantages} &
		\textbf{Application Scenarios} \\
		
		\midrule
		
		Cummins et al. \cite{40} &
		Compiler optimization framework &
		Enhances disassembly processes &
		Compiler optimization \\
		
		Hu et al. \cite{41} &
		Prompting strategies incorporating context to enhance VQA performance &
		Strengthens contextual understanding &
		Knowledge-based applications \\
		
		Hu et al. \cite{42} &
		LLM-TIKG: Threat Intelligence Knowledge Graph framework &
		Enhances cybersecurity analysis &
		Threat Intelligence Analysis \\
		
		Qian et al. \cite{43} &
		Infrastructure for large-scale LLM training &
		Efficiently handles large-scale training tasks &
		LLM training infrastructure \\
		
		Sun et al. \cite{44} &
		UniCoder: Intermediate representation to improve downstream NLP task processing &
		Enhances the model’s ability to handle complex tasks &
		Model scaling \\
		
		Zhang et al. \cite{45} &
		BB-GeoGPT: Framework tailored for Geographic Information Science &
		Highly adaptable to domain-specific knowledge &
		Geographic Information Science (GIS) \\
		
		Wang et al. \cite{46} &
		VideoAgent: Vision-language integrated system for long-form video interpretation &
		Improves multimodal reasoning capabilities &
		Multimodal reasoning \\
		
		Xia et al. \cite{47} &
		LLMGA: LLM-based generation assistant combining image generation and editing tools &
		Precise control over image synthesis, &
		Image generation and editing \\
		
		Zhou et al. \cite{48} &
		Weak-to-Strong Search: Greedy search for aligning large models with smaller counterparts &
		Compute-efficient, enhances model generalization &
		Model alignment and fine-tuning \\
		
		Land et al. \cite{49} &
		Glitch token analysis and detection method &
		Improves tokenizer robustness &
		tokenizer optimization \\
		
		\bottomrule
	\end{tabularx}
	
\end{table*}
pipeline, including knowledge extraction, fusion, representation, reasoning, retrieval, and access to a wide range of common knowledge bases \cite{50}.\par

First, the core knowledge base of LLM comes from massive text data. These text data cover a wide range of topics and fields, including books, articles, technical literature, social media content, etc. By learning these contents, LLM can master the basic rules of language, such as grammar, spelling, syntactic structure, etc., and can understand the meaning in various contexts. For example, by analyzing news, blogs, forums and other content, the model can obtain current popular topics, cultural backgrounds, and common expressions \cite{51,52}. A large amount of text training data provides LLM with a wide range of background knowledge, enabling it to generate natural and accurate language in a variety of situations. In addition to disordered text data, the knowledge base of LLM also includes\par
\begin{figure}[t]
	\centering
	\includegraphics[width=0.96\linewidth]{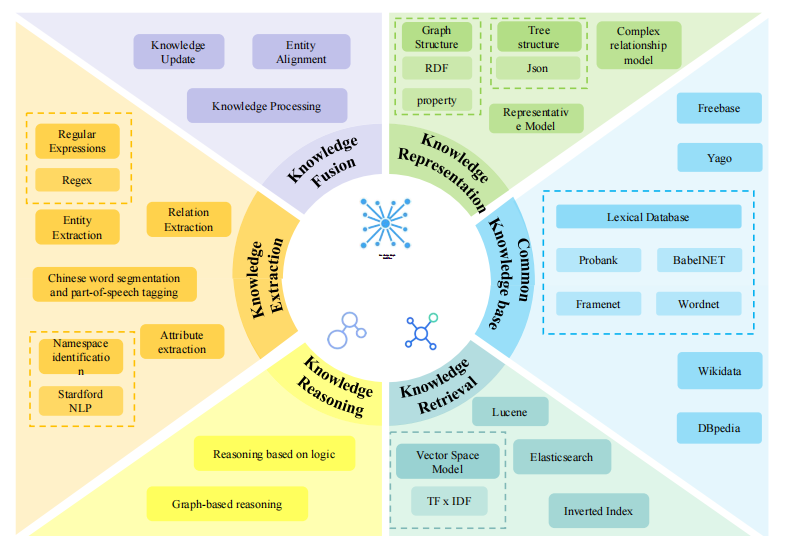}
	\caption{Flowchart of knowledge base construction and use in large language model.}
	\label{fig:figure6}
\end{figure}
structured knowledge. This part usually exists in a more systematic and standardized form, which can help the model obtain information and make inferences more accurately. For example, knowledge graphs organize entities (such as people, places, events, etc.) and their relationships, enabling models to understand complex factual relationships. In addition, domain-specific databases and tables can also provide accurate data support for models, especially when performing tasks that require accurate factual support, such as medical diagnosis and legal analysis \cite{53}.\par
The pre-training process of LLM relies on large-scale corpora, which come from public online resources such as Common Crawl, BooksCorpus, Wikipedia, etc \cite{54}. These pre-training corpora not only provide language diversity for the model, but also help the model accumulate a lot of common sense and world knowledge. Through this stage of training, LLM can answer general questions without specific context, understand cultural differences around the world, and generate natural language output that meets most situations. In order to meet the needs of specific fields, LLM can also enhance expertise through domain-specific fine-tuning. For example, medical literature, legal provisions, financial reports, etc \cite{55}. can be used as special training data sets to help models provide more accurate answers in specific fields. Through fine-tuning, LLM can generate answers that meet the standards of the field when handling specific tasks, thereby improving its professionalism and accuracy.\par

In addition, LLM also has the ability to learn interactively. The model not only learns through static training data, but also continuously updates and adjusts its knowledge base through real-time interaction with users. During the conversation with the user, the model can understand the user’s needs and adjust the response content according to the context, generating answers that are closely related to the current conversation. This interactive learning method enables LLM to continuously adapt to new information and scenarios \cite{56}. Finally, LLM also has deep cultural and language background knowledge. Because its training data covers multiple languages and cultural backgrounds, LLM can understand and generate content in different languages and cultural backgrounds. In a multilingual environment, it can recognize different expressions, idioms, and cultural differences, and generate text that conforms to local customs and language norms. This enables LLM to communicate effectively across cultures and languages in global applications.\par
LLM’s knowledge base is the core support for its powerful capabilities, covering multiple sources from massive text data to fine structured knowledge, from domain expertise to interactive learning. As technology continues to develop, LLM can show higher accuracy and flexibility in more complex tasks \cite{57}. The diversity of these knowledge bases enables LLM to not only provide support in everyday language processing tasks, but also provide precise and efficient solutions in professional fields.\par

\subsection{Reasoning ability}
After the pre-training phase, the large language model needs to be further applied to a variety of practical tasks. This phase is called the inference phase. In this phase, the model not only needs to generate language, but also needs to demonstrate its ability to understand, analyze and reason about input information to adapt to different types of downstream tasks \cite{58}.\par
To support the efficient deployment and reasoning performance of large language models in this phase, various infrastructure optimization technologies have emerged. As illustrated in Figure 7, these technologies are broadly categorized into software-level and hardware-level improvements. On the software side, distributed computing frameworks such as DeepSpeed and Megatron-LM enable large-scale parameter parallelism and pipeline parallelism, effectively reducing training and inference latency. Low-bit quantization methods—such as INT8 or FP4 compression—significantly reduce memory and computation demands, making large models more suitable for edge or mobile devices. Additionally, service-oriented deployment strategies allow model inference to be integrated into microservices or containerized environments, facilitating flexible and scalable model serving.\par
From a hardware perspective, advances in memory management (e.g., Flash-Attention and unified memory systems) and computation acceleration (e.g., tensor core optimization, dedicated AI chips like TPU/GPU) have greatly enhanced the runtime efficiency of large models. Meanwhile, new emerging technologies, such as neuromorphic computing and optical AI processors, offer potential breakthroughs for handling increasingly complex reasoning tasks at lower power and latency costs.\par
The types of reasoning tasks are rich and varied, covering open natural language generation tasks (such as dialogue systems, creative text generation, etc.) and structured and complex cognitive tasks. Although the current large language model has made significant breakthroughs in the generation ability in open scenarios, there are still certain performance bottlenecks when facing reasoning tasks with high cognitive requirements \cite{59}. Especially in the three major areas of mathematical reasoning, logical reasoning and knowledge reasoning, the model’s capabilities are still difficult to meet the requirements of high-precision applications.\par

First of all, mathematical reasoning tasks require the model to not only understand mathematical terms and symbolic language, but also have the ability to parse the problem structure, so as to propose an effective solution path and obtain the correct answer. Such tasks often involve complex numerical calculations and multi-step logical deductions, which put forward high requirements on the model’s abstract thinking ability and control of the\par
\begin{figure}[t]
	\centering
	\includegraphics[width=0.96\linewidth]{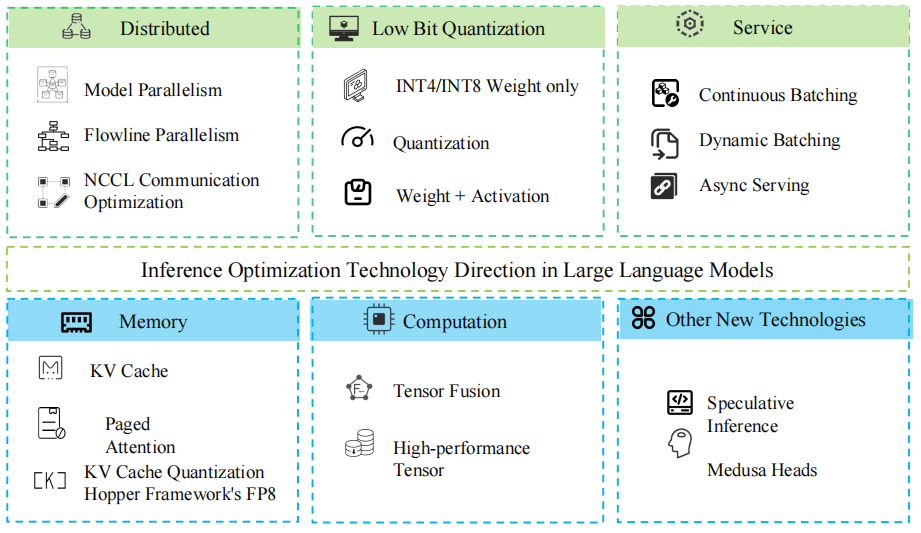}
	\caption{Summary of inference optimization techniques in large language models.}
	\label{fig:figure7}
\end{figure}
reasoning chain. However, most large language models currently have limited mathematical knowledge, and their generation process is prone to logical jumps or numerical errors, which affects the accuracy and interpretability of reasoning.\par
Secondly, logical reasoning emphasizes the use of rules, propositional relationships and deductive reasoning methods from the perspective of formal logic to systematically analyze and judge problems. This type of task usually involves complex processes such as multi-jump reasoning, conditional selection, and proposition combination, which poses a challenge to the logical consistency and deductive integrity of the model. Although large language models can learn certain pattern recognition capabilities through big data training, it is often difficult to achieve human-level accurate judgment when dealing with highly structured or strictly logical content \cite{60}.\par
Finally, knowledge reasoning tasks focus on whether the model can complete problem solving based on background knowledge or common sense. The key to this type of task lies in the understanding and integration of factual information. However, since the current large model training process mainly relies on statistical learning of massive corpora rather than building a systematic knowledge system, the model is prone to "knowledge hallucination" during the generation process, that is, generating information that seems reasonable but is actually wrong. The root of this problem is that the model’s grasp of knowledge is probabilistic and fragmentary, rather than having the ability to deeply understand semantics and integrate knowledge logic.\par

In summary, although large language models have demonstrated strong capabilities in language generation, the tasks they face in the reasoning stage are often more complex and more structured, especially in key areas such as mathematics, logic, and knowledge reasoning, where there is still significant room for improvement. Based on this, this article will focus on the above three types of reasoning tasks, aiming to explore effective methods and strategies to further improve the performance and practicality of large language models in complex reasoning scenarios.\par

\subsection{Embodied Intelligence}

\subsubsection{Introduction to Embodied Intelligence}
The term embodied refers to an intelligent agent possessing a physical body that enables both sensing and motor control. This body is not merely a structural form but serves as the foundation for the agent’s interaction with the real world \cite{61}. It allows the agent to influence its environment through actions and to perceive the resulting changes, thereby achieving closed-loop learning and adaptive behavior. An embodied system is not only physically present in space but can also actively engage with its surroundings—identifying objects, manipulating tools, navigating obstacles—thus establishing "physicality" as the key to genuine autonomous intelligence \cite{62}.\par
To further classify and understand the landscape of embodied intelligence, Figure 8 provides a structured taxonomy that includes key dimensions such as Swarm Embodied Intelligence, Embodied Entity, Embodied Perception, Embodied Evolution, Embodied Common Sense, and Embodied Cognition. This taxonomy reflects the multi-faceted nature of embodied AI. Swarm Embodied Intelligence emphasizes the collective behavior of multiple agents, inspired by biological systems like ants or bees, where complex global behaviors emerge from local interactions. Embodied Entity focuses on the physical manifestation of the agent, including its hardware and design architecture \cite{63}. Embodied Perception highlights the agent’s ability to interpret multimodal signals (e.g., visual, auditory, tactile), while Embodied Evolution refers to the system’s capacity to iteratively improve its control and reasoning mechanisms through interaction and adaptation. Additionally, Embodied Common Sense and Embodied Cognition reflect the system’s internal representation of the world and its understanding of physical affordances, causality, and action-outcome relationships.\par
\begin{figure}[t]
	\centering
	\includegraphics[width=0.96\linewidth]{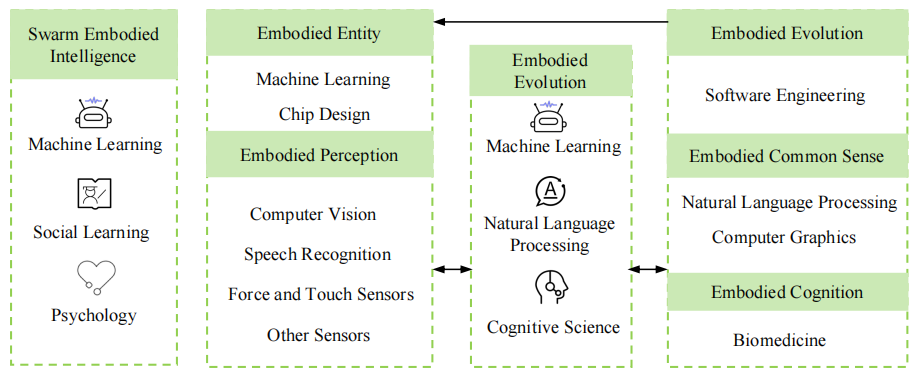}
	\caption{The core components and development path of embodied intelligence.}
	\label{fig:figure8}
\end{figure}

Embodied AI refers to artificial intelligence systems that combine cognitive capabilities with a physical body, allowing them to interact dynamically with the environment. Unlike traditional AI models trained solely on static, pre-collected data, embodied AI emphasizes learning through real-time interaction. Representative examples include service robots, autonomous vehicles, and bionic robots. These systems exhibit human-like physical actions and continuously refine their strategies based on observation, action, and feedback, enabling them to handle complex, real-world tasks \cite{64}. Specifically, an embodied intelligent agent must be capable of: (1) understanding human language and instructions, (2) decomposing high-level commands into specific subtasks, (3) perceiving and localizing target objects in the environment, (4) performing physical actions such as movement and manipulation, and\par
(5) adapting its behavior in response to environmental feedback. In contrast, disembodied AI refers to mainstream models such as large language models and image recognition systems, which typically rely on pre-labeled, static datasets curated by humans. These models lack the capability for physical interaction with the real world and thus represent a form of intelligence rooted more in symbolic reasoning or virtual manipulation. While effective in tasks like language understanding and image generation within virtual environments, disembodied AI tends to exhibit limited generalizability in dynamic, real-world contexts and lacks the autonomy and adaptability inherent to embodied systems. Embodied intelligent robots are the most tangible representation of embodied AI, equipped with the abilities to perceive, reason, act, and learn \cite{65}. They typically feature:\par
(1) multimodal perception, utilizing cameras, microphones, and tactile sensors to gather rich sensory data including images, sound, temperature, and vibration; (2) semantic understanding and reasoning, enabling them to interpret complex commands and make decisions based on environmental cues; (3) mobility and manipulation, allowing for spatial navigation, obstacle avoidance, and physical interaction with objects; (4) interactive learning, through which they acquire new knowledge via environmental engagement and continuously optimize their behavior. These robots can perform not only repetitive tasks—such as vacuum cleaning—but also more sophisticated roles in household assistance, medical support, warehouse logistics, and disaster response, thus evolving into truly "smart and useful" physical intelligences. Embodied tasks refer to those that require agents to actively observe, move, communicate, and interact in a physical setting, akin to human behavior \cite{66}. These tasks demand not only robust perceptual and cognitive capabilities but also adaptive responsiveness to complex and dynamic environments. Examples include interactive instruction or cooperative tasks, such as assisting the elderly with medication or assembling equipment in collaboration with humans. These scenarios are typically characterized by environmental uncertainty, ambiguous goals, and intricate feedback mechanisms, serving as critical benchmarks for evaluating the practical competence of embodied intelligent systems.\par

\subsubsection{The development of embodied intelligence}
The development of embodied intelligence can be divided into five key stages, which fully reflect the evolution of artificial intelligence from a reasoning paradigm based on symbolic logic to a multimodal intelligent system driven by a large model that integrates perception, language and action \cite{67}. This process not only reflects the continuous breakthroughs in technology, but also shows the profound impact of the cross-disciplinary integration of cognitive science, neuroscience and robotics on the research paradigm of artificial intelligence. The development process of embodied intelligence is shown in Figure 9.\par
1. Origin Stage (1940–1990): The early phase was primarily influenced by Cybernetics, Computationalism, and Symbolic AI. Research during this period focused on logic-based reasoning and rule systems \cite{68}. Despite certain achievements in formal representation, this paradigm gradually faced criticism due to its lack of modeling for environmental interaction and embodied perception.\par
\begin{figure}[t]
	\centering
	\includegraphics[width=0.96\linewidth]{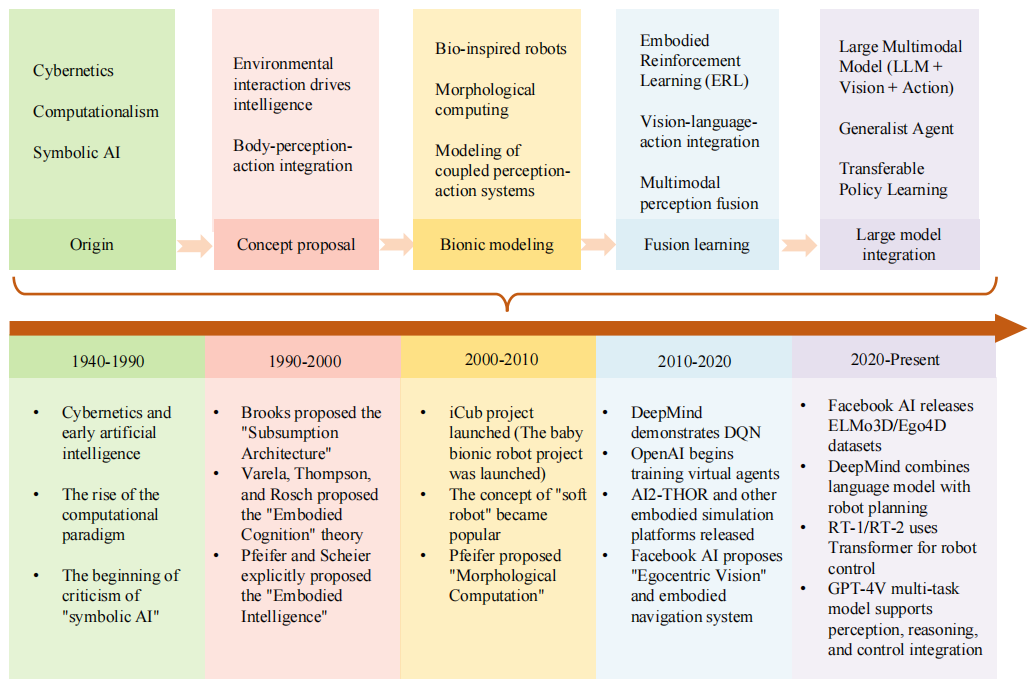}
	\caption{The development of embodied intelligence.}
	\label{fig:figure9}
\end{figure}
2. Conceptual Proposal Stage (1990–2000): Brooks proposed the “Subsumption Architecture,” emphasizing decentralized control. Varela, Thompson, and Rosch introduced the theory of “Embodied Cognition,” while Pfeifer and Scheier explicitly formulated the concept of “Embodied Intelligence,” laying the foundation for integrated body-perception-action intelligence \cite{69}.\par
3. Bionic Modeling Stage (2000–2010): Research in embodied intelligence transitioned into experimental implementations. Projects such as iCub and “soft robots” gained popularity. Pfeifer proposed the idea of “Morphological Computation,” highlighting the computational role of physical morphology in intelligent behavior.\par
4. Fusion Learning Stage (2010–2020): Multimodal integration and environmental interaction became core focuses. DeepMind demonstrated DQN in reinforcement learning; OpenAI began training virtual agents \cite{70}; simulation platforms like AI2-THOR emerged. Facebook AI proposed the concept of “Egocentric Vision” and developed embodied navigation systems, promoting the fusion of vision, language, and action.\par
5. Large Model Integration Stage (2020–Present): The forefront of research has shifted towards generalist agents that integrate large language models (LLMs) with visual and action modules. Facebook AI released datasets such as ELMo3D and Ego4D \cite{71}. DeepMind combined language models with robot planning. OpenAI introduced RT-1/RT-2 series, which use Transformer architectures for robotic control. GPT-4V represents a multimodal model supporting perception, reasoning, and action in a unified loop.\par

Through active interaction with the physical world, embodied agents demonstrate strong adaptability in a variety of forms . Among them, robots are the most typical and most mature\par
\begin{figure}[t]
	\centering
	\includegraphics[width=0.96\linewidth]{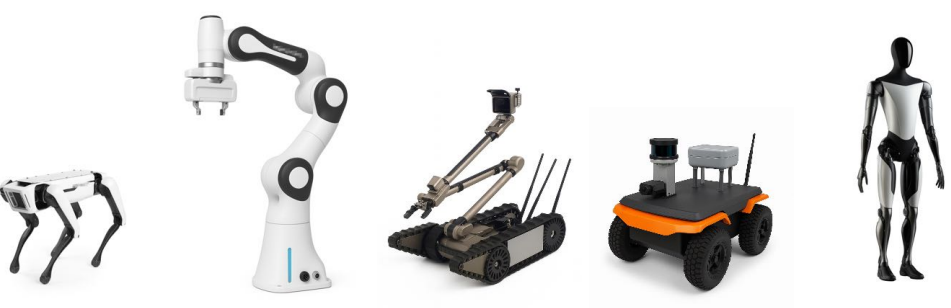}
	\caption{Different categories of Embodied Robots: (a) Quadruped Robots, (b) Fixed-base Robots, (c) Tracked Robots, (d) Wheeled Robots, and (e) Humanoid Robots.}
	\label{fig:figure10}
\end{figure}
embodied form. Depending on the task scenario and operational requirements, robots show significant differences in structural design and execution methods, as shown in Figure 10.\par
(a) Quadruped robots Inspired by animals in nature, quadruped robots achieve gait flexibility and terrain adaptability through multi-joint design, can maintain stable walking on complex surfaces, and are often used for rescue, reconnaissance and exploration tasks. Representative platforms include Boston Dynamics’ Spot \cite{72}. These systems have excellent dynamic balance and environmental perception capabilities, and support remote control and autonomous navigation. However, their high manufacturing costs and limited battery life still restrict large-scale promotion, especially in long-term missions or outdoor scenes \cite{73}.\par
(b) Fixed-base robots Fixed-base robots are widely used in industrial manufacturing, education and scientific research due to their stable structure and high precision \cite{74}. They are usually equipped with high-resolution sensors and precision actuators, capable of performing micron-level operations, and are suitable for tasks with high repeatability and strict precision requirements. For example, representative platforms such as Franka Emika Panda, KUKA iiwa and Sawyer all have good programming adaptability. However, the fixed-position design limits its workspace and mobility, making it difficult to meet the needs of collaborative scenarios or large-scale deployment \cite{75}.\par
(c) Tracked Robots Tracked robots have excellent off-road performance and terrain adaptability due to their stable track structure \cite{76}. They are widely used in tasks in high-risk or complex environments such as post-disaster rescue, military reconnaissance, and building inspection \cite{77}. Compared with wheeled robots, tracked platforms have stronger traction and passability in rugged, soft, and inclined terrains. Their larger contact area can effectively disperse pressure and reduce the risk of getting stuck in mud or sand. A typical example is iRobot’s PackBot, which has modular robotic arms and multi-sensor fusion capabilities, and can perform reconnaissance, explosive ordnance disposal, and humanitarian rescue tasks under remote control. However, the track system itself has some limitations, such as energy efficiency loss caused by relatively high mechanical friction, and the movement speed on flat ground is not as flexible as that of wheeled platforms.\par
(d) Wheeled and tracked robots Mobile robots have higher environmental adaptability and are widely used in logistics, security, agriculture and disaster response \cite{78}. Wheeled robots are known for their simple structure, low energy consumption, and strong mobility on flat ground. Typical representatives include Kiva and Jackal, which are often equipped with lidar and cameras\par

to support autonomous navigation. However, their performance is limited in rugged terrain \cite{79}.\par
(e) Humanoid robots Humanoid robots have human-like structures and movement capabilities, and are suitable for performing service, collaboration and auxiliary tasks \cite{80}. Their dexterous hands and high-degree-of-freedom joint systems enable them to perform precision operations, such as medical assistance, guided tours, etc. Typical examples include Boston Dynamics’ Atlas, Honda’s ASIMO, and AIST’s HRP series \cite{81}. By integrating large language models (LLMs), humanoid robots are expected to achieve more intelligent task responses and human-machine interactions in the future.\par

\subsubsection{Embodied intelligence perception, decision-making and control execution}
As an important research direction in the artificial intelligence system, embodied intelligent robots have achieved rapid development in both basic theory and practical engineering application in recent years \cite{82}. Its core concept is to closely integrate perception, cognition and action generation, so that robots can effectively interact and complete tasks in real physical environments. In particular, in the neural-symbolic integrated design framework, the intelligent agent builds a technical closed loop through the four stages of "perception-reasoning-execution-feedback" to support efficient, intelligent and adjustable task response.\par
1. The perception of embodied intelligence The perception module is the first step for the robot to interact with the outside world. It provides basic information support for cognitive reasoning and control \cite{83}. Typical embodied intelligent robots integrate multiple types of sensor systems and can obtain environmental data from multiple dimensions such as vision, force perception, and touch, thereby achieving more accurate judgment and execution.\par
(1) Visual information acquisition The visual perception system enables the robot to obtain an understanding of the target scene from image or video data \cite{84}. In the structural disassembly scenario, the robot uses image recognition algorithms to extract the contour, estimate the size and locate the target object in space. By deploying visual models such as convolutional neural networks, the system can automatically identify different part structures and assist in generating operation path planning, effectively improving its task adaptability \cite{85}.\par
(2) Force feedback mechanism Force perception is mainly used to capture the physical force characteristics generated by the robot during operation. In the intelligent disassembly task, the force sensor monitors the torque change when contacting the object in real time, so that the system can adjust the operation strategy according to the feedback, such as reducing the force to avoid damaging fragile parts, or increasing the traction to complete the removal of key connection points. This process ensures the unity of operation accuracy and equipment safety.\par
(3) Surface tactile recognition Tactile perception gives the robot the ability to distinguish the surface details of the object, such as material type, surface roughness or temperature changes. In actual operation, the tactile module helps the robot determine the material characteristics of the target object and choose the most appropriate grasping or operation method accordingly. For example, a rough surface may require a stronger clamping force, while a smooth surface needs to avoid the risk of slipping or sliding off.\par
2. Decision-making with Embodied Intelligence In an embodied intelligent system, the decision-making module is like a "brain" that is responsible for integrating sensory data, planning task processes, and making intelligent\par

responses \cite{86}. With the rapid development of large language models (LLMs) in the field of reasoning and language understanding, their role in embodied intelligent decision-making systems has become increasingly prominent. In particular, in the context of collaborative modeling with neural symbolic systems, LLMs, as a "language-driven reasoning engine", can empower robots with stronger contextual understanding and task adaptation capabilities \cite{87}.\par
(1) Collaborative mechanism of neural learning, symbolic reasoning, and LLM Traditional neural networks can efficiently extract pattern features when processing sensory inputs (such as images, point clouds, etc.), while symbolic systems provide a formal expression of task logic structures and operation processes \cite{88}. By introducing LLMs, robots can bridge sensory results with symbolic information - for example, converting image recognition results into natural language descriptions, and then LLM generates operation strategies based on prompts. In addition, LLMs can act as "language intermediaries" to understand task instructions, infer the order of operations, and assist symbolic reasoning modules in making decisions under rule constraints, such as identifying conditional logic such as "remove screws first and then disconnect cables".\par
(2) Language-guided adaptive decision-making for dynamic scenarios In unstructured environments, traditional decision-making systems are difficult to respond flexibly to emergencies. LLM’s contextual understanding and language reasoning capabilities provide robots with a new path to dynamically adjust their strategies \cite{89}. When the robot encounters an unexpected situation (such as rusted parts or structural obstructions), it can generate a scene description through visual perception and use LLM to generate a response plan, such as "try to clean the surface and then re-operate" or "replace with a tool with stronger torque." This "language-assisted dynamic decision-making mechanism" significantly improves the robot’s ability to respond to complex scenarios and interactivity.\par
(3) Learning optimization and instruction fine-tuning: LLM-driven strategy evolution The autonomous learning ability of embodied agents can also be continuously enhanced with the help of LLM \cite{90}. Based on traditional reinforcement learning or imitation learning, LLM can summarize the cause of failure by parsing task feedback language (such as "operation failed because the tool slipped") and reversely optimize the strategy template. At the same time, with the help of instruction fine-tuning or reinforcement learning enhanced language models (such as RLHF), LLM can continuously optimize its strategy generation process, enabling the robot to have stronger generalization and adaptability when facing new tasks \cite{91,92}.\par
3. Execution of Embodied Intelligence As the link closest to the physical world in the embodied system, the execution module is the bridge from strategy generation to action implementation \cite{93}. As the application of LLM in embodied tasks continues to expand, its auxiliary capabilities in the execution stage are gradually emerging, especially in motion interpretation, operation semantic transformation and tool matching, showing the potential of language cognition to guide physical behavior.\par
(1) Language-driven path control and interpretation In complex scenarios, robots need to perform path control operations with semantic layer guidance. With the help of LLM, the system can not only plan obstacle avoidance paths, but also map low-level control signals into semantic paths that can be interpreted by humans (such as "move 5cm to the left to avoid obstacles, then grab the parts"). This coupling of the language layer and the motion layer not only enhances the interpretability of human-machine collaboration, but also makes it easier for the system to integrate instruction fine-tuning or feedback training processes.\par

(2) Language mapping and switching suggestions for tool operations The end effector module faces a variety of operation scenarios (such as rotating screws and separating connectors), and often needs to quickly switch between different tools \cite{94}. LLM can generate tool suggestions based on the current task semantics, object attributes and historical experience, and assist the execution module to automatically select the optimal actuator. For example, "The identified part is a transparent plastic shell. It is recommended to use a soft suction cup instead of the gripper." This type of language prompt can significantly improve the accuracy of tool use and scene adaptability.\par

\subsubsection{Analysis of Difficulties of Embodied Intelligence}
One of the foremost challenges in realizing embodied intelligence is the construction of a robust, universal ontology platform with strong mobility and manipulation capabilities \cite{95}. Overcoming the bottlenecks in key hardware technologies and developing robot products that are reliable, cost-effective, and highly versatile remains a formidable task. In the pursuit of "general-purpose intelligence," humanoid robots are widely regarded as one of the ultimate forms of embodied agents \cite{96}. As a result, research and development focused on humanoid structures continues to be a critical and hotly debated topic across both academia and industry.\par

Another major challenge lies in designing intelligent agent systems equipped with advanced cognitive functions \cite{97}. As the core component of embodied intelligence, these systems must not only respond effectively to complex and dynamic real-world environments but also exhibit a range of sophisticated cognitive and interactive abilities. Specifically, intelligent agents are expected to demonstrate: (1) precise perception of three-dimensional physical environments; (2) robust task scheduling, execution, and adaptive adjustment capabilities; (3) comprehensive general knowledge and multi-level semantic reasoning; (4) natural and coherent human-machine interaction, particularly in multi-turn dialogues; (5) construction and utilization of long-term memory mechanisms; (6) personalized services and emotional sensitivity; and (7) strong generalization across tasks, along with self-learning and transfer learning capabilities. To achieve such functionality, agents must be capable of real-time perception and decision-making, which imposes stringent requirements on the speed and efficiency of data collection, transmission, and processing. Meanwhile, current large language models (LLMs) demand substantial computational resources, making it extremely difficult to deploy them within resource-constrained embedded robotic systems while maintaining low-latency responses for high-complexity reasoning and decision-making tasks \cite{98}. As illustrated in the figure 11, the embodied intelligence system operates through a perception–cognition–action loop that mirrors human-like reasoning in real-world environments. Specifically, the system first perceives multimodal inputs from the environment, including visual, auditory, and tactile information. This sensory data is then abstracted into high-level representations and passed to a central "brain" module, where LLMs are leveraged for memory retrieval, knowledge learning, inference, and decision-making. The agent analyzes contextual cues such as sky conditions and online forecasts to infer weather outcomes. Drawing upon its internal knowledge base and prior memory, the system makes a decision—e.g., advising to carry an umbrella—and initiates a robotic action to deliver the tool accordingly. This highlights the potential of combining LLMs with embodied perception and actuation mechanisms, enabling intelligent agents to understand user instructions, reason about environmental conditions, and execute goal-oriented behaviors.\par
However, implementing such an integrated framework on physical agents remains challenging due to the high computation and memory demands of LLMs, especially under the constraints of real-time interaction. Bridging this gap will require innovations in model compression, edge computing, and hierarchical reasoning architectures.\par
A further obstacle is the lack of high-quality industry data, which significantly hampers the training of embodied intelligent systems \cite{99}. Real-world environments are inherently complex and continuously evolving, yet current data resources are often insufficient in terms of diversity, richness, and contextual relevance. Particularly in mission-critical applications where task success rates must be exceptionally high, reliance on broad, generic datasets is inadequate. Embodied intelligence is characterized by a high degree of coupling with the physical world, meaning that meaningful and effective data can only be acquired through deployment in real environments—unlike traditional disembodied AI systems that depend on pre-collected datasets \cite{100}. Therefore, the construction of high-quality, domain-specific datasets becomes especially vital in key sectors. A practical approach to mitigate this challenge is to adopt a hierarchical design for agent systems and constrain task execution within defined scenarios, thereby striking a balance between enhancing generalization and ensuring high task success rates.\par

Finally, a fundamental aspect of embodied intelligence is its capacity for continuous evolution, which depends on the agent’s ability to learn autonomously and improve over time. In adapting to new environments, agents that are better aligned with their morphological characteristics tend to acquire effective problem-solving strategies more rapidly. However, due to the near-infinite space of possible agent morphologies, it is impractical to explore all design permutations within limited computational resources. Moreover, the degrees of freedom inherent in the physical design of the ontology can impose constraints on task adaptability and learning capacity, ultimately affecting the performance of the controller in learning and decision-making. There exists a complex and as-yet-unresolved interplay among environmental complexity, agent morphology, and learning ability. Addressing how to achieve rapid, efficient strategy learning and rational decision-making under resource constraints represents a critical frontier in the future advancement of embodied intelligence.\par
\begin{figure}[t]
	\centering
	\includegraphics[width=0.96\linewidth]{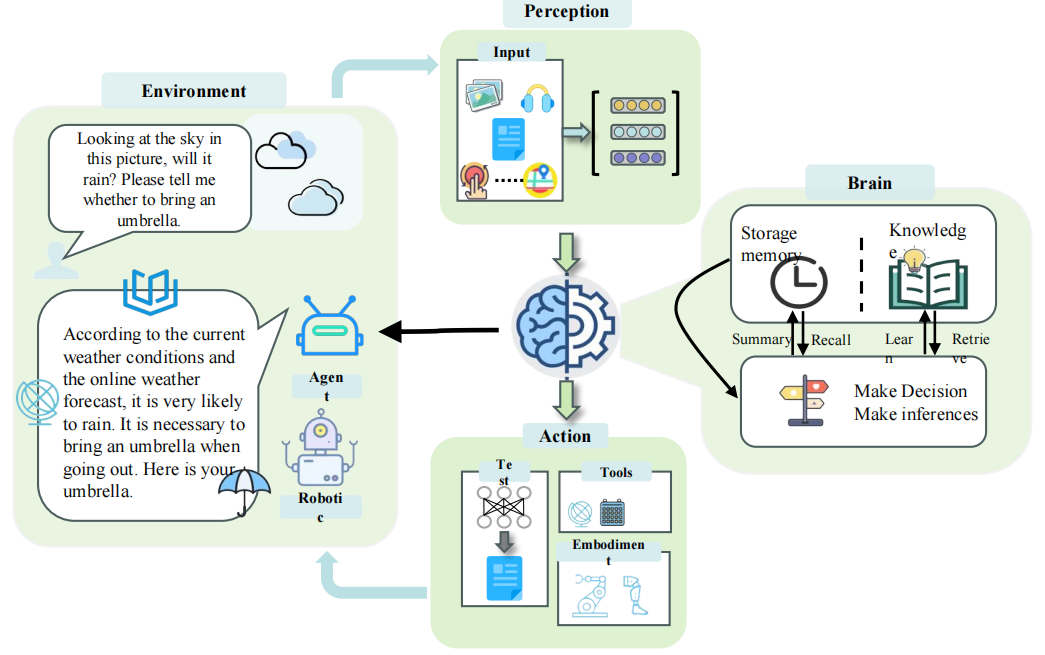}
	\caption{Example diagram of the integration of embodied intelligence and large language model.}
	\label{fig:figure11}
\end{figure}
\subsubsection{Latest Research on Embodied Intelligence}
Embodied intelligence represents a multifaceted paradigm that integrates physical morphology, sensing, actuation, and adaptive control within autonomous systems. Zardini et al. \cite{101} propose a structured co-design methodology rooted in monotone co-design theory, which supports the interdisciplinary integration of heterogeneous engineering domains to address the inherent complexity of embodied intelligence systems. Their framework emphasizes the use of analytical and simulation-based modeling techniques to enable cohesive and systematic design solutions .\par
Advancements in soft robotics have further demonstrated the feasibility of embodied intelligence through innovations in materials and actuation mechanisms. Ma et al. \cite{102} introduce fully soft actuators that integrate sensing, actuation, and control at the single-unit level, presenting a simplified yet autonomous strategy for constructing soft robotic systems with embodied capabilities . In parallel, Mengaldo et al. \cite{103} offer a comprehensive analysis of the physical principles underpinning embodied intelligence in soft robotics, underscoring the critical role of physical interaction modeling in enhancing system performance.\par
Learning and evolutionary strategies have played a central role in advancing embodied intelligence. Gupta et al. \cite{104} present Deep Evolutionary Reinforcement Learning (DERL), a framework that evolves diverse agent morphologies to solve complex locomotion and manipulation tasks. Their work highlights the interplay between environmental complexity, morphological adaptability, and control learnability, reinforcing the notion that embodied intelligence benefits significantly from adaptive learning mechanisms grounded in environmental interaction.\par
The integration of machine learning within embodied systems has also been critically examined. Roy et al. \cite{105} argue that embodied intelligence should not be viewed merely as a target application for machine learning, but rather as a conceptual foundation that challenges and reshapes traditional learning paradigms. Their perspective emphasizes the need for learning frameworks that account for the physical embodiment and energy-information exchanges with the environment. Extending this view, Iida et al. \cite{106} investigate the role of temporal dynamics, illustrating how self-organization and emergent behavior in autonomous systems are fundamentally influenced by multi-scale temporal interactions.\par
Recent efforts have explored embodied intelligence within human-in-the-loop learning frameworks. Long et al. \cite{107} demonstrate the utility of interactive simulation environments in training surgical robots, illustrating how human input can enhance embodied learning and support the development of robust control policies through simulation-based training.\par

The convergence of embodied intelligence and emerging AI technologies, particularly large language models (LLMs), is a rapidly evolving area of research. Zeng et al. \cite{108} provide a survey on the application of LLMs in robotics, highlighting their potential to augment robot perception, decision-making, and human-robot interaction. Building upon this, Fan et al. \cite{109} explore the use of LLMs in industrial robotics, enabling more autonomous and adaptive manufacturing systems. Feng et al. \cite{110} further extend this trajectory by examining LLM-driven spatial intelligence across multiple domains, including embodied agents, smart cities, and geospatial systems, thereby demonstrating the interdisciplinary impact of these models .\par
In summary, the current body of literature underscores that embodied intelligence is an inherently interdisciplinary and evolving field. It brings together physical design, adaptive learning, physics-based modeling, and cutting-edge AI methodologies. The integration of these dimensions holds the promise of significantly advancing the autonomy, adaptability, and versatility of intelligent systems across a wide range of real-world applications.\par
\begin{table*}[t]
	\centering
	\caption{Summary of Key Contributions in Embodied Intelligence Research}
	\label{tab:embodied_intelligence_research}
	
	\small
	\renewcommand{\arraystretch}{1.08}
	\setlength{\tabcolsep}{4pt}
	
	\begin{tabularx}{\textwidth}{
			>{\raggedright\arraybackslash}p{2.1cm}
			>{\centering\arraybackslash}p{1.1cm}
			>{\raggedright\arraybackslash}p{3.2cm}
			>{\raggedright\arraybackslash}X
			>{\raggedright\arraybackslash}p{3.0cm}
		}
		\toprule
		
		\textbf{Author} &
		\textbf{Year} &
		\textbf{Research Topic} &
		\textbf{Key Contribution} &
		\textbf{Application Domain} \\
		
		\midrule
		
		Zardini et al. \cite{101} &
		2020 &
		Co-Design Framework &
		Structured co-design for integration &
		General systems \\
		
		Ma et al. \cite{102} &
		2021 &
		Soft Robotics Actuators &
		Integrated sensing \& actuation &
		Soft robotics \\
		
		Gupta et al. \cite{104} &
		2021 &
		DERL for Adaptation &
		Co-evolution of body and control &
		Locomotion \& control \\
		
		Roy et al. \cite{105} &
		2021 &
		ML in Embodied Systems &
		Energy-aware physical learning &
		Robotics \& ML \\
		
		Mengaldo et al. \cite{103} &
		2022 &
		Physics Modeling &
		Physics-based modeling &
		Soft robotics \\
		
		Iida et al. \cite{106} &
		2022 &
		Temporal Dynamics &
		Timescale-based self-organization &
		Adaptive systems \\
		
		Long et al. \cite{107} &
		2023 &
		Human-in-the-loop Learning &
		Interactive simulation for training &
		Surgical robotics \\
		
		Zeng et al. \cite{108} &
		2023 &
		LLMs for Robotics &
		Survey of LLM-enhanced robotics &
		Human-robot interaction \\
		
		Fan et al. \cite{109} &
		2024 &
		LLMs in Industry &
		LLM-based autonomous control &
		Industrial robots \\
		
		Feng et al. \cite{110} &
		2025 &
		Spatial AI Reasoning &
		Spatial AI via LLMs &
		Multiscale AI systems \\
		
		\bottomrule
	\end{tabularx}
	
\end{table*}

Moreover, interdisciplinary efforts are increasingly investigating spatial intelligence powered by LLMs across a range of scales—from individual embodied agents to smart cities and geospatial systems. These developments reveal the potential of LLMs to enhance spatial reasoning and situational awareness, inspiring new directions in embodied intelligence research \cite{110}. In summary, embodied intelligence is an inherently multidisciplinary field that integrates physical design, learning mechanisms, physics-based modeling, temporal dynamics, and advanced computational tools such as LLMs. Ongoing research highlights its centrality in developing autonomo\par

\section{Current state of technology integration}

With the continuous advancement of technology, the field of artificial intelligence is undergoing a profound change. In this context, language models (LLMs), as one of the most representative technologies at present, have gradually been integrated into a variety of intelligent systems, promoting the development of various innovative applications. In this process, the integration of LLMs with knowledge bases, reasoning models, and embodied intelligence is particularly prominent, showing great potential and application value. As illustrated in the figure 12, the modern intelligent agent system is structured into four major functional layers: the representation layer, the knowledge layer, the reasoning layer, and the action layer (Embodied Intelligence). Large language models, such as GPT-4, serve as the core of the representation layer, utilizing transformer-based architectures, self-supervised learning, in-context reasoning, and multimodal fusion capabilities to perform natural language understanding and generation.\par
\begin{figure}[t]
	\centering
	\includegraphics[width=0.90\linewidth]{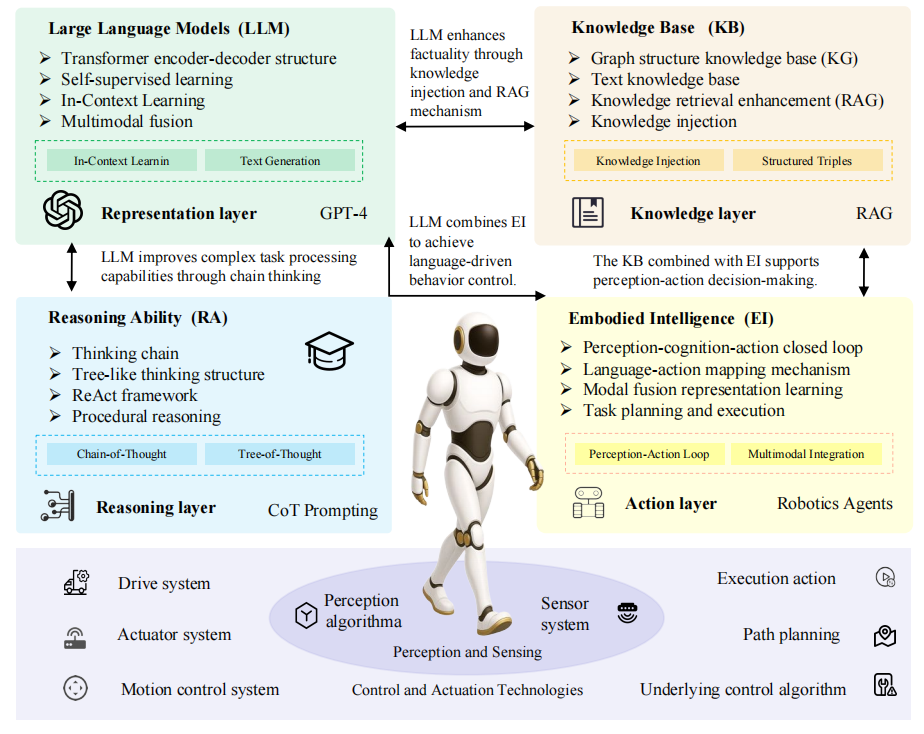}
	\caption{Embodied Intelligence, Large Language Model, Knowledge Base and Reasoning Ability together constitute the core architecture of the intelligent system.}
	\label{fig:figure12}
\end{figure}

The knowledge layer incorporates structured graph-based knowledge bases (KGs) and textual repositories, enhanced through retrieval-augmented generation (RAG) and dynamic knowledge injection. This enables LLMs to retrieve, interpret, and incorporate factual information more effectively. The reasoning layer further builds on this foundation by introducing structured inference methods such as Chain-of-Thought (CoT), Tree-of-Thought, the ReAct framework, and procedural reasoning strategies, empowering agents with the ability to decompose problems and formulate step-by-step decisions. At the highest level of physical intelligence lies the action layer, where Embodied Intelligence (EI) bridges abstract language reasoning with concrete environmental interaction. EI achieves this through a perception-cognition-action closed loop, language-to-action mapping, and multimodal sensory fusion. Through coordinated modules including sensor systems, perception algorithms, motion control, and execution logic, embodied agents (e.g., robots) can not only understand natural language instructions but also perceive their surroundings and respond with appropriate physical actions.\par
This hierarchical integration of LLMs with knowledge, reasoning, and embodiment components demonstrates a promising architecture for next-generation intelligent systems. It allows AI agents to not only "think and speak", but also to "see, decide, and act"—bringing us closer to truly autonomous, interactive, and adaptive artificial general intelligence.

\subsection{LLM and Knowledge Base}
When discussing the integration of LLM and knowledge base, it is crucial to understand how the two interact with each other. The powerful language generation capability of LLM provides a new perspective for reasoning and querying of knowledge graphs, while knowledge graphs provide structured factual support and background information for LLM. On this basis, the ability of LLM is not limited to generating language texts, but can also be further applied to task processing in specific fields, forming a deep integration of knowledge base and LLM, thereby promoting the development of domain-specific applications.

\subsubsection{LLM and Knowledge Graph Integration}
Daniel Adam and Tomáš Kliegr \cite{111} introduced a method for verifying RDF triples using LLMs, where RDF statements are validated by comparing them to external documents, achieving a precision of 88\%. This approach avoids relying on internal LLM factual knowledge but requires human oversight. Yichun Feng et al. \cite{112} proposed the Knowledge Graph-based Thought (KGT) framework, which integrates verified information from knowledge graphs to improve LLM responses and reduce factual errors, particularly in the biomedical field. Xinbang Dai et al. \cite{113} empirically studied how LLMs process knowledge graph information in different input formats, showing that linearized triples are more effective for answering fact-intensive questions, though results vary across LLMs. John S. Erickson et al. \cite{114} enhanced LLM workflows through Retrieval-Augmented Generation (RAG) and knowledge graphs, improving traceability and explainability, though the integration with external tools remains challenging. Xiang Shi et al. \cite{115} proposed an LLM-based generative retrieval framework that enhances relevance and trustworthiness in web search results, outperforming traditional methods in certain query types. Juan Sequeda et al. \cite{116} highlighted the role of knowledge graphs in supporting LLM-powered enterprise question-answering systems, providing a formal framework for query validation and result explanation, although scalability remains an issue for complex systems.\par

\subsubsection{LLM for Domain-Specific Applications}

FeiLong Wang et al. \cite{117} introduced the LLM-KGMQA system, a multi-hop question-answering framework for the medical field, which significantly improves accuracy in entity linking and reasoning, achieving 99.80\% accuracy, but requires substantial computational resources. Jingchi Jiang et al. \cite{118} developed the knowledge-guided agricultural LLM (KALLM) for agricultural decision-making, which outperforms existing frameworks by enhancing query precision, although data scarcity in the agricultural domain is a limitation. Wissal Benjira et al. \cite{119} presented an LLM-augmented knowledge graph approach for mapping open data to Sustainable Development Goals (SDGs), achieving high accuracy in mapping, but relying on the availability of diverse metadata. Samuel Kernan Freire et al. \cite{120} proposed an LLM-powered knowledge-sharing system for manufacturing, designed\par
\begin{table*}[t]
	\centering
	
	\small
	\renewcommand{\arraystretch}{1.08}
	\setlength{\tabcolsep}{4pt}
		\caption{LLM and Knowledge Graph Integration}
	\label{tab:llm_kg_integration}
	\begin{tabularx}{\textwidth}{
			>{\raggedright\arraybackslash}p{2.3cm}
			>{\raggedright\arraybackslash}X
			>{\raggedright\arraybackslash}X
			>{\raggedright\arraybackslash}X
		}
		\toprule
		
		\textbf{Author(s)} &
		\textbf{Innovation} &
		\textbf{Advantages} &
		\textbf{Drawbacks} \\
		
		\midrule
		
		Daniel et al. \cite{111} &
		Verifying RDF triples using LLMs by comparing RDF statements to external documents &
		High precision (88\%) in RDF verification &
		Requires human oversight, method may be slower \\
		
		Feng et al. \cite{112} &
		Knowledge Graph--based thought framework to improve LLM responses with verified KG information &
		Reduces factual errors, enhances drug--cancer associations &
		Limited to biomedical, requires cross-domain testing \\
		
		Dai et al. \cite{113} &
		Empirical study on LLMs' comprehension of KG input formats and prompt organization &
		Improves LLM performance in KG-based tasks &
		Varying effectiveness across different models, noisy subgraphs \\
		
		Erickson et al. \cite{114} &
		Enhancing LLM workflows with RAG and KGs to ensure traceability and flexibility &
		More accurate and explainable LLM outputs &
		Integration with external tools can be complex \\
		
		Xiang et al. \cite{115} &
		LLM-based generative retrieval framework focusing on trustworthy search results &
		Improves relevance, responsibility, and trustworthiness &
		Works better with certain query types, not universally applicable \\
		
		Sequeda et al. \cite{116} &
		LLM-powered enterprise question answering, supported by KGs &
		Provides formal validation and trusted data for queries &
		May not scale well for large or complex queries in diverse enterprises \\
		
		\bottomrule
	\end{tabularx}

\end{table*}
to retrieve information from factory documentation, which facilitates quicker information retrieval but still faces a preference for human experts in real-world settings. Tao Xie et al. \cite{121} introduced a theme-based lecture summary system using LLMs and graph-based theme segmentation, which enhances knowledge recall and student satisfaction, though its effectiveness is dependent on the structure of the lecture material. Lizhuang Sun et al. \cite{122} proposed SF-GPT, a training-free method for enhancing LLM-based knowledge graph construction, improving knowledge recall and F1 scores, but may face challenges with certain datasets and complex knowledge fusion tasks.\par

\subsection{LLM and Reasoning Model}

Wu et al. \cite{123} proposed a Long-to-Short (L2S) reasoning framework and combined it with model merging technology to solve the "overthinking" phenomenon of current large language models in deep logical reasoning. This method merges models through multiple methods such as task vectors, singular value decomposition (SVD), activation information, etc., and achieves significant shortening of the reasoning path (average response length is reduced by 55\%) while maintaining or even improving the performance of the original model on multiple complex tasks. Experiments have shown that this method is stable in models of different sizes (1.5B to 32B), and has self-examination and adaptive\par
\begin{table*}[t]
	\centering
	
	\small
	\renewcommand{\arraystretch}{1.08}
	\setlength{\tabcolsep}{4pt}
	\caption{LLM for Domain-Specific Applications}
	\label{tab:llm_domain_specific}
	\begin{tabularx}{\textwidth}{
			>{\raggedright\arraybackslash}p{2.2cm}
			>{\raggedright\arraybackslash}X
			>{\raggedright\arraybackslash}X
			>{\raggedright\arraybackslash}X
		}
		\toprule
		
		\textbf{Authors} &
		\textbf{Innovation} &
		\textbf{Advantages} &
		\textbf{Drawbacks} \\
		
		\midrule
		
		Wang et al. \cite{117} &
		LLM-KGMQA: Multi-hop question-answering system for the medical field, addressing entity linking and reasoning &
		Improved accuracy (99.80\%) and robust performance &
		Requires significant computational resources for entity linking and multi-hop reasoning \\
		
		Jiang et al. \cite{118} &
		Knowledge-guided agricultural LLM for domain-specific decision-making &
		State-of-the-art performance in agriculture, improves accuracy in agricultural queries &
		Data scarcity in the domain, challenges in precision retrieval \\
		
		Benjira et al. \cite{119} &
		LLM-augmented KG for SDG indicator mapping, integrating open data &
		High precision in mapping open data to SDGs &
		Requires access to high-quality and diverse metadata \\
		
		Freire et al. \cite{120} &
		LLM-powered knowledge-sharing system for manufacturing, retrieving information from factory documentation &
		Enhances quick information retrieval and resolution of issues &
		Preference for human experts still prevails, open-source models may have privacy concerns \\
		
		Xie et al. \cite{121} &
		Theme-based lecture summary system using LLMs and graph-based segmentation &
		Enhances knowledge recall and student satisfaction &
		Effectiveness varies with lecture structure \\
		
		Sun et al. \cite{122} &
		SF-GPT: A training-free method for enhancing LLM-based KG construction &
		Improved recall and F1 score, no training required &
		May not work well with all types of datasets or knowledge fusion \\
		
		\bottomrule
	\end{tabularx}

\end{table*}

response capabilities, demonstrating the good balance between efficiency and depth of L2S reasoning. Tung et al. \cite{124} proposed the GreenMind-Medium-14B-R1 model for low-resource scenarios in Vietnamese, and combined the Group Relative Policy Optimization strategy with the language detection mechanism to fine-tune Vietnamese synthetic reasoning data. Its innovation lies in the design of a dual reward function to alleviate the problem of language mixing and improve the factuality of the reasoning content. Experiments on the VLSP 2023 and SeaExam multilingual test sets show that GreenMind not only outperforms existing methods in task accuracy, but also achieves significant improvements in language consistency, verifying its potential in the field of low-resource language reasoning. Ma et al. \cite{125} proposed the SQL-R1 model for the NL2SQL reasoning task, breaking through the previous limitation of relying solely on supervised fine-tuning (SFT) training, and instead used reinforcement learning to design a special reward mechanism to adapt to complex semantic structures such as multi-table joins and nested queries. They further used small-scale synthetic data to complete the training, and combined with data engineering optimization strategies, they finally achieved 88.6\% and 66.6\% execution accuracy on the Spider and BIRD datasets, respectively, indicating that the method has good generalization performance even in the case of insufficient data. Costas et al. \cite{126} proposed the GNN-RAG framework, which systematically integrated graph neural networks (GNNs) and large language models into the retrieval-augmented generation (RAG) process for the first time. This method uses GNN to reason about entities and answer candidates from dense subgraphs and construct the shortest path in the graph to achieve explicit knowledge chain expression. These paths are processed by language and then input into LLM to complete the question answering task. Experimental results show that this method achieves the current best performance on both WebQSP and CWQ datasets, especially in multi-hop, multi-entity question answering tasks, which is 8.9–15.5 percentage points higher than the existing methods. Sui et al. \cite{127} proposed the FiDeLiS framework, which designs a training-free method combining deductive paths and step-by-step verification to address the hallucination problem in knowledge graph enhanced question answering. The framework uses beam search and deductive scoring mechanism to control the reasoning path, and combines the Path-RAG strategy to narrow the search space to reduce the computational burden. It achieves higher accuracy and interpretability on multiple KGQA datasets, and is particularly suitable for improving the factual consistency of the reasoning process. Luo et al. \cite{128} focused on the reasoning ability of LLM in recommendation systems and proposed the RALLRec+ framework to jointly model representation learning and chain reasoning. This method integrates the project descriptions and collaborative signals generated by the language model in the retrieval stage, and introduces knowledge enhancement prompts and multi-model fusion mechanisms in the generation stage. By dynamically sorting to adapt to changes in user preferences, the experiment showed excellent results in three real data sets, verifying its practical value in recommendation reasoning.\par

Yang et al. \cite{129} proposed the Reasoning-based Bias Detector (RBD) module to improve the consistency and fairness of large model evaluation. The module works as an external plug-in, detects judgment bias in the generated results through structured reasoning, and performs multiple rounds of self-correction. The study constructed an evaluation dataset containing four types of bias (redundancy, position, conformity, and emotion), and trained the RBD model at multiple parameter scales. The results show that RBD improves the evaluation accuracy by 18.5\% and the consistency by 10.9\% on eight LLM evaluators on average, far exceeding the baseline method. Zhang et al. \cite{130} focus on the reasoning ability in multimodal document understanding and propose the DocAssistant model, which combines data augmentation with the MLLM data generator to generate step-by-step question and answer data, significantly improving the model’s ability to handle complex structures and multi-hop question and answer. Its experimental results on the InfoVQA and ChartQA datasets show that the performance is improved by 5\% and 7\% respectively, proving the effectiveness of step-by-step reasoning in document understanding tasks. Beary et al. \cite{131} studied the text and image reasoning capabilities of large language models in radiation oncology and proposed a hybrid approach that combines fine-tuning and hint engineering. By fine-tuning the Llama 3.2 model with QLoRA and integrating hinting technology, the reasoning performance was significantly improved on a newly constructed challenging medical dataset, demonstrating the strong adaptability and scalability of this method for medical tasks. Tan et al. \cite{132} proposed the Hydra framework, which integrates multi-source evidence from knowledge graphs and documents in a structured manner to support LLM for multi-hop and multi-entity reasoning tasks. The framework introduces a ternary verification mechanism to jointly evaluate the credibility, semantic consistency, and path structure of the data source. On seven datasets, Hydra\par
\begin{table*}[t]
	\centering
	\caption{Category 1: Research Based on Reasoning Frameworks and Methods}
	\label{tab:reasoning_frameworks_methods}
	
	\small
	\renewcommand{\arraystretch}{1.08}
	\setlength{\tabcolsep}{3pt}
	
	\begin{tabularx}{\textwidth}{
			>{\raggedright\arraybackslash}p{1.8cm}
			>{\centering\arraybackslash}p{1.0cm}
			>{\raggedright\arraybackslash}X
			>{\raggedright\arraybackslash}X
			>{\raggedright\arraybackslash}X
		}
		\toprule
		
		\textbf{Authors} &
		\textbf{Year} &
		\textbf{Innovation} &
		\textbf{Advantages} &
		\textbf{Limitations} \\
		
		\midrule
		
		Wu et al. \cite{123} &
		2025 &
		Proposed Long-to-Short (L2S) reasoning and model merging. &
		Increases reasoning efficiency while maintaining performance. &
		Merging efficiency depends on model size, adds computational complexity. \\
		
		Tung et al. \cite{124} &
		2025 &
		Introduced GreenMind, using strategy optimization for reasoning. &
		Outperforms previous works on Vietnamese reasoning tasks. &
		Limited to specific languages, struggles with low-resource languages. \\
		
		Ma et al. \cite{125} &
		2025 &
		Introduced SQL-R1, a reinforcement learning-based model for NL2SQL. &
		Improves accuracy in complex database tasks. &
		Reinforcement learning training may suffer from cold start issues. \\
		
		Costas \cite{126} &
		2024 &
		GNN-RAG combines GNN reasoning with LLMs in a retrieval-augmented generation style. &
		Optimizes graph data processing, improves knowledge graph QA. &
		Performance depends on graph data quality, limited in complex reasoning. \\
		
		Sui et al. \cite{127} &
		2024 &
		Proposed FiDeLiS for factuality-enhanced LLM reasoning. &
		Improves factuality and interpretability. &
		Dependency on reasoning steps increases computational cost. \\
		
		Luo et al. \cite{128} &
		2025 &
		Introduced RALLRec+ for enhancing recommendation system reasoning. &
		Enhances recommendation reasoning, handles dynamic user preferences. &
		Requires continuous data updates, may still be less effective in real-time applications. \\
		
		\bottomrule
	\end{tabularx}
	
\end{table*}

outperformed the strong baseline ToG-2 by an average of 20.3\%, and enabled the medium-sized model Llama-3.1-8B to achieve the reasoning performance of GPT-4-Turbo. Kunat et al. \cite{133} focused on improving the reasoning ability of low-resource languages LLM and proposed to migrate the reasoning ability of DeepSeek R1 to language-specific models through model merging. Taking Thai LLM as an example, this study demonstrated that only using public data and about \$120 of computing resources can enhance its logical reasoning ability while maintaining the performance of the original language, providing a feasible path for localized LLM reasoning. Lan et al. \cite{134} proposed LLM4QA, which converts natural language questions into SPARQL queries through instruction fine-tuning to achieve efficient reasoning on structured knowledge graphs. This method combines chain thinking generation with unsupervised entity relationship retrieval, significantly improving the reasoning accuracy and query efficiency of the model, and has important application prospects in knowledge-driven question-answering systems. Li et al. \cite{135} proposed the LINKED framework, which integrates knowledge extraction, filtering and integration modules, focusing on improving the accuracy of LLM in common sense reasoning tasks. By introducing a reward model to remove noisy knowledge and designing a consistent reasoning mechanism, it effectively reduces invalid reasoning. It performs well on two common sense reasoning benchmarks and innovatively proposes the "validity retention score" indicator. Inoue et al. \cite{136} proposed the DrugAgent system for drug target prediction tasks, introducing a multi-agent architecture into the reasoning process. Each agent in the system integrates knowledge graphs, literature evidence, and machine learning predictions, and uses Chain-of-Thought and ReAct strategies to achieve transparent reasoning. On the kinase inhibitor dataset, the F1 score increased by 45\%, and an explainable biomedical reasoning path was provided, which has important clinical decision-making value. Phuc et al. \cite{137} proposed the QUERY2TREE model, which combines GNN and Google Gemini series models to embed entity description information, build a set of logical operations based on K-D trees, and support complex queries such as projection, intersection, union, and negation. On the FB15k, FB15k-237, and NELL995 datasets, QUERY2TREE surpasses multiple baseline models in MRR and hits@3 metrics, with maximum improvements of 69.6\% and 69.4\%, respectively, significantly verifying its effectiveness in knowledge graph logical reasoning.\par

\subsection{LLM and Embodied Intelligence}
With the rapid evolution of Large Language Models (LLMs), their integration into embodied systems has enabled new frontiers in robot perception, decision-making, and real-world interaction. LLMs empower agents with the ability to reason, plan, and learn from natural language instructions, greatly enhancing the autonomy and versatility of embodied intelligence. This section systematically reviews the latest research on LLM-driven embodied intelligence, categorized into four main themes: task execution and planning, multimodal enhancement, multi-agent collaboration, and security robustness.\par

\subsubsection{Embodied task execution and autonomous planning driven by LLM}
One of the most active research areas focuses on how LLMs can be used to guide embodied agents in performing complex tasks. These studies explore how LLMs interpret instructions, decompose long-horizon tasks, and generate executable action sequences in dynamic environments. The ability of LLMs to understand context and reason over sequences plays a pivotal role in enabling embodied agents to autonomously navigate, manipulate, and interact with the physical world.\par

In the field of LLM-driven embodied task execution and autonomous planning, researchers have explored diverse approaches leveraging the capabilities of large language models. Haolin Fan et al. \cite{138} proposed a three-stage framework for industrial robotic task execution using LLMs, where GPT-4 demonstrated excellent performance in complex manufacturing scenarios, though it remains limited in handling 3D spatial tasks. Yaran Chen et al. \cite{139} developed RoboGPT, a hierarchical planning system capable of decomposing long-horizon tasks and adapting to environmental feedback, albeit with high dependency on large-scale datasets and multi-module integration. Ruaridh Mon-Williams et al. \cite{149} introduced the ELLMER framework, combining retrieval-augmented generation with multimodal feedback for adaptive planning; however, its evaluation remains restricted to domestic scenarios. Luxi Li et al. \cite{140} explored the application of multimodal LLMs to autonomous mining vehicles, presenting a promising concept that lacks detailed implementation. Chan Hee Song et al. \cite{150} proposed LLM-Planner for few-shot planning, achieving competitive performance under minimal supervision, though its reliance on environmental modeling remains a constraint. Andrew Szot et al. \cite{141} presented a unified\par
\begin{table*}[t]
	\centering
	\caption{Category 2: Application-Specific Reasoning Research}
	\label{tab:application_specific_reasoning}
	
	\small
	\renewcommand{\arraystretch}{1.08}
	\setlength{\tabcolsep}{3pt}
	
	\begin{tabularx}{\textwidth}{
			>{\raggedright\arraybackslash}p{1.8cm}
			>{\centering\arraybackslash}p{1.0cm}
			>{\raggedright\arraybackslash}X
			>{\raggedright\arraybackslash}X
			>{\raggedright\arraybackslash}X
		}
		\toprule
		
		\textbf{Authors} &
		\textbf{Year} &
		\textbf{Innovation} &
		\textbf{Advantages} &
		\textbf{Limitations} \\
		
		\midrule
		
		Yang et al. \cite{129} &
		2025 &
		Introduced Reasoning-based Bias Detector (RBD). &
		Reduces bias in evaluations, improves consistency. &
		Needs significant supervision for bias detection. \\
		
		Zhang et al. \cite{130} &
		2024 &
		Designed step-wise multimodal models for document reasoning. &
		Enhances multimodal understanding, improves complex question answering. &
		Relies heavily on high-quality labeled data. \\
		
		Beary et al. \cite{131} &
		2025 &
		Introduced hybrid fine-tuning to enhance medical domain reasoning. &
		Improves reasoning accuracy in radiation oncology. &
		Domain-specific, limited transferability to other medical fields. \\
		
		Tan et al. \cite{132} &
		2025 &
		Proposed Hydra framework for cross-source enhanced reasoning. &
		Excels in multi-hop and multi-entity reasoning tasks. &
		Requires diverse data sources, computationally intensive. \\
		
		Kunat et al. \cite{133} &
		2025 &
		Merged language-specific LLMs with advanced reasoning models. &
		Enhances reasoning in low-resource languages. &
		Can be computationally expensive for merging models. \\
		
		Lan et al. \cite{134} &
		2024 &
		Proposed LLM4QA for graph-based reasoning with SPARQL queries. &
		Improves knowledge graph reasoning efficiency. &
		Relies on structured knowledge graphs, limited in unstructured data. \\
		
		Li et al. \cite{135} &
		2024 &
		Proposed LINKED for filtering and integrating knowledge in reasoning tasks. &
		Improves commonsense reasoning accuracy. &
		Performance may degrade with noisy knowledge sources. \\
		
		Inoue et al. \cite{136} &
		2025 &
		Introduced DrugAgent, a multi-agent LLM system for drug-target prediction. &
		Enhances reliability and transparency in drug-target predictions. &
		Multi-agent approach may face challenges in information integration. \\
		
		Phuc et al. \cite{137} &
		2025 &
		Proposed QUERY2TREE for reasoning over knowledge graphs. &
		Improves logical query accuracy with knowledge graph embeddings. &
		Dependent on knowledge graph quality, may struggle with complex queries. \\
		
		\bottomrule
	\end{tabularx}
	
\end{table*}

Generalist Embodied Agent (GEA), which generalizes across diverse tasks but requires complex training procedures. Vishnu Dorbala et al. \cite{151} developed LGX, achieving state-of-the-art results in zero-shot object navigation, though tested primarily in simulation. R. Schumann et al. \cite{152} proposed VELMA, which transforms environmental vision into\par
\begin{table*}[t]
	\centering
	\caption{LLM-driven Embodied Task Execution and Planning}
	\label{tab:llm_embodied_task_execution}
	
	\small
	\renewcommand{\arraystretch}{1.08}
	\setlength{\tabcolsep}{3pt}
	
	\begin{tabularx}{\textwidth}{
			>{\raggedright\arraybackslash}p{2.1cm}
			>{\centering\arraybackslash}p{1.0cm}
			>{\raggedright\arraybackslash}X
			>{\raggedright\arraybackslash}X
			>{\raggedright\arraybackslash}X
		}
		\toprule
		
		\textbf{Authors} &
		\textbf{Year} &
		\textbf{Innovation} &
		\textbf{Advantages} &
		\textbf{Limitations} \\
		
		\midrule
		
		Fan et al. \cite{138} &
		2025 &
		Proposed a three-stage framework for industrial robot task execution using LLMs &
		Outstanding GPT-4 performance in industrial tasks &
		Limited ability in 3D spatial planning \\
		
		Chen et al. \cite{139} &
		2025 &
		Designed RoboGPT with long-term decomposition and feedback adjustment &
		Adaptive subgoal updating &
		Requires large datasets and complex modules \\
		
		Ruaridh et al. \cite{140} &
		2025 &
		Introduced ELLMER with RAG for multi-step reasoning &
		Strong sensorimotor adaptation &
		Tested mainly in home-like tasks \\
		
		Li et al. \cite{141} &
		2024 &
		Proposed multi-modal LLMs for autonomous mining driving &
		Applicable to realistic mining scenarios &
		Conceptual work, lacks implementation \\
		
		Song et al. \cite{142} &
		2023 &
		Introduced LLM-Planner for few-shot planning &
		Strong performance with limited data &
		Weak environmental state modeling \\
		
		Andrew et al. \cite{143} &
		2025 &
		Unified GEA model across embodied tasks via MLLM &
		High cross-domain generalization &
		Data-hungry and complex training \\
		
		Vishnu et al. \cite{144} &
		2023 &
		Proposed LGX for zero-shot navigation &
		Large improvement in success rate &
		Validated mostly in simulation \\
		
		Schumann et al. \cite{145} &
		2024 &
		VELMA verbalizes visual context for navigation &
		Greatly improves city-scale VLN &
		Requires explicit visual-language mapping \\
		
		Wu et al. \cite{146} &
		2023 &
		TaPA plans executable actions with scene grounding &
		Outperforms GPT-3.5 in planning &
		Object recognition impacts execution \\
		
		Danny et al. \cite{147} &
		2023 &
		PaLM-E integrates vision, state, and language &
		Versatile across modalities and tasks &
		Costly training and deployment \\
		
		Mower et al. \cite{148} &
		2024 &
		ROS-LLM enables non-expert robot programming via chat &
		Supports multiple behavior modes &
		Relies on full ROS ecosystem \\
		
		\bottomrule
	\end{tabularx}
	
\end{table*}

contextual prompts for real-world navigation, requiring explicit visual-language mapping. Zhenyu Wu et al. \cite{153} introduced the TaPA framework for grounded planning based on scene perception, showing improvements in executable planning, while object detection accuracy remains a bottleneck. Danny Driess et al. \cite{142} built PaLM-E, a multimodal model for robotic reasoning tasks, which performs well across modalities but incurs high computational costs. Finally, Christopher Mower et al. (2024) presented ROS-LLM, enabling natural language programming for robots by non-experts through ROS integration, though it depends heavily on the ROS ecosystem.\par

\subsubsection{Embodied intelligence system with multimodal fusion and perception enhancement}
To achieve robust embodied intelligence, it is essential to go beyond language alone. This subsection highlights systems that integrate LLMs with multimodal sensory inputs, such as vision, audio, and environmental sensors. These systems enhance perception, situation awareness, and interaction fluency. Through multimodal grounding, LLMs become more capable of interpreting real-world scenes and adjusting behavior in a context-sensitive manner.\par
\begin{table*}[t]
	\centering
	\caption{Multi-Modal Fusion and Perception Enhancement in Embodied Systems}
	\label{tab:multimodal_fusion_perception}
	
	\small
	\renewcommand{\arraystretch}{1.08}
	\setlength{\tabcolsep}{3pt}
	
	\begin{tabularx}{\textwidth}{
			>{\raggedright\arraybackslash}p{2.1cm}
			>{\centering\arraybackslash}p{1.0cm}
			>{\raggedright\arraybackslash}X
			>{\raggedright\arraybackslash}X
			>{\raggedright\arraybackslash}X
		}
		\toprule
		
		\textbf{Authors} &
		\textbf{Year} &
		\textbf{Innovation} &
		\textbf{Advantages} &
		\textbf{Limitations} \\
		
		\midrule
		
		Sun et al. \cite{149} &
		2025 &
		Multi-level LLM framework for coal mine sensor analysis &
		Efficient risk prediction and learning &
		Requires simulated and expert knowledge \\
		
		Abhay et al. \cite{154} &
		2024 &
		EnvGen: LLM-generated adaptive RL training environments &
		Boosts training efficiency &
		Depends on feedback quality \\
		
		Zheng et al. \cite{155} &
		2023 &
		Steve-Eye combines LLM and visual encoders &
		Enables end-to-end multimodal interaction &
		Requires large datasets \\
		
		Ma et al. \cite{156} &
		2025 &
		Perspective on LLMs in autonomous driving &
		Offers future research outlook &
		Lacks technical solutions \\
		
		Liu et al. \cite{157} &
		2024 &
		EAI-SIM simulation platform with ROS &
		Controls UAVs and arms in photo-realistic sim &
		Needs powerful simulation hardware \\
		
		Song et al. \cite{158} &
		2024 &
		Guide-LLM aids visually impaired via text maps &
		Combines commonsense and path planning &
		Small sample evaluation \\
		
		Micol et al. \cite{159} &
		2025 &
		VITA for adaptive mental well-being coaching &
		Personalized multi-modal robot coaching &
		Limited experimental scale \\
		
		Kolby et al. \cite{160} &
		2023 &
		DECKARD agent with Dream/Wake exploration &
		Greatly improves RL efficiency &
		Relies on LLM hypothesized subgoals \\
		
		Yang et al. \cite{161} &
		2024 &
		EMMA distilled from text-world LLM to visual world &
		Successfully transfers to vision tasks &
		Needs dual-modality synchronization \\
		
		\bottomrule
	\end{tabularx}
	
\end{table*}

In the domain of multimodal fusion and perception-enhanced embodied systems, researchers have explored how to integrate language, vision, and sensor data to improve LLM-driven perception and decision-making. Yi Sun et al. \cite{149} constructed a multi-level LLM architecture for coal mine safety assessment using heterogeneous sensor inputs, significantly enhancing evaluation efficiency, though it relies on simulation environments and prior knowledge bases. Abhay Zala et al. \cite{154} proposed EnvGen, a framework that generates training environments to boost RL agent performance, improving training efficiency while being sensitive to feedback precision. Sipeng Zheng et al. \cite{155} introduced Steve-Eye, an end-to-end multimodal model enabling perception in open-world settings, but it demands extensive datasets. Yunsheng Ma et al. \cite{156} provided a prospective review on multimodal LLMs in autonomous driving, highlighting key future challenges and directions. Guocai Liu et al. \cite{157} developed EAI-SIM, a simulation platform integrating ROS and Isaac Sim to facilitate embodied LLM task evaluation, although requiring high computational resources. Sangmim Song et al. \cite{158} proposed Guide-LLM for indoor navigation assistance for visually impaired users, demonstrating effective reasoning with limited evaluation scale. Micol Spitale et al. \cite{159} developed VITA, a multimodal LLM-based well-being coach, showing positive user outcomes but limited by small sample size. Kolby Nottingham et al. \cite{160} introduced DECKARD, a two-phase exploration framework using LLM-guided abstract world modeling for RL tasks, significantly improving sample efficiency while being sensitive to initial modeling accuracy. Yijun Yang et al. \cite{161} presented EMMA, which distills LLM behavior from a text-based world into a vision-based agent, achieving superior generalization but requiring complex cross-modal synchronization.\par

\subsubsection{Multi-agent collaboration and organizational strategy}
Beyond individual intelligence, embodied agents are increasingly expected to collaborate in teams. This section introduces works that explore the organizational structures and communication strategies that govern multi-agent collaboration. These studies often draw inspiration from human team dynamics and propose structured prompts, role assignments, and adaptive learning mechanisms to improve coordination and efficiency among LLM-based agents.\par
\begin{table*}[t]
	\centering
	\caption{Multi-Agent Collaboration and Organizational Strategies}
	\label{tab:multi_agent_collaboration}
	
	\small
	\renewcommand{\arraystretch}{1.08}
	\setlength{\tabcolsep}{3pt}
	
	\begin{tabularx}{\textwidth}{
			>{\raggedright\arraybackslash}p{2.2cm}
			>{\centering\arraybackslash}p{1.0cm}
			>{\raggedright\arraybackslash}X
			>{\raggedright\arraybackslash}X
			>{\raggedright\arraybackslash}X
		}
		\toprule
		
		\textbf{Authors} &
		\textbf{Year} &
		\textbf{Innovation} &
		\textbf{Advantages} &
		\textbf{Limitations} \\
		
		\midrule
		
		Zhang et al. \cite{162} &
		2024 &
		ReAd mechanism improves multi-agent LLM efficiency &
		Fewer LLM queries and agent steps &
		Evaluation mainly in simulation \\
		
		Liu et al. \cite{163} &
		2024 &
		FaGeL uses smart fabric for implicit feedback &
		Learns user preference non-intrusively &
		High system complexity \\
		
		Guo et al. \cite{164} &
		2024 &
		Organizational prompting reduces LLM agent conflict &
		Boosts teamwork and efficiency &
		Validation in real-world pending \\
		
		\bottomrule
	\end{tabularx}
	
\end{table*}

In the field of multi-agent collaboration and organizational strategies, researchers have investigated how LLM-based agents coordinate instructions and optimize strategies in cooperative settings. Yang Zhang et al. \cite{162} proposed the ReAd framework, which uses advantage regression learning to improve collaborative agent performance, significantly reducing LLM query costs, though currently validated only in simulations. Jia Liu et al. \cite{163} introduced FaGeL, an embodied agent that leverages smart fabric sensors for implicit human-agent interaction, offering adaptive learning capabilities but with high system complexity. Xudong Guo et al. \cite{164} explored prompt-based organizational structures to reduce communication overhead among LLM agents, effectively improving team efficiency, although broader validation in real-world scenarios remains limited.\par

\subsubsection{Security and robustness challenges of embodied intelligence and LLM}
As LLMs gain control over physical agents, the issue of safety and reliability becomes critical. This subsection reviews the emerging threats and challenges that arise from adversarial prompts, contextual backdoors, and behavioral misalignment. It also discusses current efforts in evaluation frameworks, robustness benchmarks, and attack mitigation strategies aimed at ensuring the trustworthy deployment of LLM-empowered embodied systems.\par

\begin{table*}[t]
	\centering
	\caption{Security and Robustness Challenges in Embodied LLMs}
	\label{tab:security_robustness_embodied_llms}
	
	\small
	\renewcommand{\arraystretch}{1.08}
	\setlength{\tabcolsep}{3pt}
	
	\begin{tabularx}{\textwidth}{
			>{\raggedright\arraybackslash}p{2.1cm}
			>{\centering\arraybackslash}p{1.0cm}
			>{\raggedright\arraybackslash}X
			>{\raggedright\arraybackslash}X
			>{\raggedright\arraybackslash}X
		}
		\toprule
		
		\textbf{Authors} &
		\textbf{Year} &
		\textbf{Innovation} &
		\textbf{Advantages} &
		\textbf{Limitations} \\
		
		\midrule
		
		Liu et al. \cite{150} &
		2025 &
		Contextual backdoor attacks in LLM-based agents &
		Dual-modal triggers for covert actions &
		No defense strategy yet in real use \\
		
		Zhang et al. \cite{151} &
		2024 &
		Investigated LLM-based robot jailbreak threats &
		Systematic exposure of jailbreak scenarios &
		No concrete safety mechanisms \\
		
		Li et al. \cite{152} &
		2024 &
		Universal benchmarking interface for LLM agents &
		Fine-grained error taxonomy &
		No coverage of real multimodal scenes \\
		
		Kovalev et al. \cite{153} &
		2022 &
		Reviewed LLM-based instruction planners without training &
		Works in zero-shot scenarios &
		Lack of real task validations \\
		
		Szot et al. \cite{165} &
		2023 &
		LLaRP: RL-trained LLM policy with vision &
		Superior multi-task performance &
		Needs intensive training and environment \\
		
		Leon et al. \cite{166} &
		2024 &
		Studied virtual agent personality effects in VR &
		Extrovert agents yield better experience &
		No embodiment in physical robots \\
		
		Liu et al. \cite{167} &
		2024 &
		EIRAD dataset and BLIP2 for adversarial attack test &
		Reveals attack success in LLM agents &
		Evaluation mainly on textual input \\
		
		\bottomrule
	\end{tabularx}
	
\end{table*}

In the area of security and robustness challenges, researchers have highlighted potential vulnerabilities of LLM-powered embodied agents. Aishan Liu et al. \cite{150} revealed contextual backdoor attacks that exploit poisoned demonstrations to induce latent programmatic defects, proposing dual-modality triggers while lacking effective defense strategies. H. Zhang et al. \cite{151} identified critical risks of LLM-enabled embodied AI violating safety constraints, particularly emphasizing jailbreaking risks and the misalignment between language and action spaces. Manling Li et al. \cite{152} developed the Embodied Agent Interface for fine-grained performance analysis, improving diagnostic resolution but with limited applicability to real-world scenarios. A. K. Kovalev et al. \cite{153} surveyed instruction-to-action methods for embodied AI using pretrained LLMs, reducing training demands but lacking empirical validation. A. Szot et al. \cite{165} proposed LLaRP, a reinforcement-learning-enhanced policy framework based on LLMs, achieving strong performance but requiring extensive environment interaction. Leon Kroczek et al. \cite{166} investigated how virtual LLM agents with varying personas affect user interaction in VR, revealing significant social-cognitive effects, albeit without physical embodiment. Shuyuan Liu et al. \cite{167} introduced the EIRAD dataset and novel adversarial attack strategies for robustness testing, providing actionable insights into LLM-based decision-level weaknesses, though the evaluation focus remains largely text-centric.\par

\subsubsection{Large language model drives embodied intelligent control execution}
Recent advancements in large language models (LLMs) have significantly influenced the evolution of embodied intelligence, particularly in enhancing robots’ capacity for autonomous perception, decision-making, and control execution. A growing body of research has demonstrated the potential of integrating LLMs into robotic systems, leading to novel frameworks that address the challenges of complex and dynamic real-world tasks.\par
Mon-Williams et al. \cite{168} introduced ELLMER, a GPT-4-enabled embodied LLM framework that incorporates retrieval-augmented generation for task planning in unpredictable environments. Their system successfully integrates force and visual feedback to complete multi-step tasks such as coffee preparation, showcasing LLMs’ potential for adaptive control in long-horizon robotic operations. In the domain of industrial robotics, Fan et al. \cite{169} proposed a three-stage LLM-agent-based framework that autonomously interprets process constraints, generates tool paths, and simulates embodied behavior. Experimental results highlight the superior performance of GPT-4 in structured industrial tasks, especially in high-level planning scenarios, affirming its utility in automation systems. Huang et al. \cite{170} formulated the "Code as Policies" paradigm, where code-generation-capable LLMs are repurposed to synthesize robot control policies. By leveraging few-shot prompting, the system produces executable policy code from natural language commands, enabling spatial reasoning and behavior generalization. This study validates the feasibility of using LLMs for low-level embodied control synthesis. Sun et al. \cite{171} examined the role of LLMs within Industry 5.0’s human-centric cyber-physical-social systems (CPSSs). Their review emphasized LLMs’ ability to support single-agent and multi-agent collaboration through embodied perception, multimodal instruction following, and interactive scheduling, thus laying a foundation for scalable intelligent manufacturing ecosystems. Zhao et al. \cite{172} developed MultiBotGPT, a layered control architecture for coordinating UAVs and UGVs using GPT-3.5. The system demonstrates high accuracy in task execution and significantly reduces operator workload, suggesting LLMs can facilitate intuitive multi-robot collaboration and mission control. Shen et al. \cite{173} conceptualized embodied intelligence as the bridge between virtual models and physical interaction. They underscored the necessity of grounding LLMs within physical agents to enable true general intelligence, thus motivating the shift from language-only models toward sensorimotor-coupled reasoning systems. To address multi-agent coordination in embodied scenarios, Jiang et al. \cite{174} introduced the KoMA framework, which integrates LLMs with modules for shared memory, reflective ranking, and multi-step planning. This design empowers autonomous driving agents with stronger generalization and adaptability, particularly in uncertain or emergent environments\par

Despite the notable progress made by existing studies in integrating large language models (LLMs) into embodied intelligence systems, several limitations remain. Firstly,\par
\begin{table*}[t]
	\centering
	\caption{Summary of Studies on LLM-driven Embodied Control}
	\label{tab:llm_embodied_control}
	
	\small
	\renewcommand{\arraystretch}{1.08}
	\setlength{\tabcolsep}{3pt}
	
	\begin{tabularx}{\textwidth}{
			>{\raggedright\arraybackslash}p{2.1cm}
			>{\centering\arraybackslash}p{1.0cm}
			>{\raggedright\arraybackslash}X
			>{\raggedright\arraybackslash}X
			>{\raggedright\arraybackslash}X
		}
		\toprule
		
		\textbf{Authors} &
		\textbf{Year} &
		\textbf{Innovation} &
		\textbf{Advantages} &
		\textbf{Limitations} \\
		
		\midrule
		
		Ruaridh et al. \cite{168} &
		2025 &
		Proposed ELLMER framework combining GPT-4 with RAG for task planning in unpredictable environments. &
		Enables long-horizon task execution via visual/force feedback; adaptable planning. &
		Generalization scope not fully validated. \\
		
		Fan et al. \cite{169} &
		2024 &
		Introduced LLM agents in manufacturing for autonomous design, planning, and control. &
		Achieved 81.88\% task completion with GPT-4 in complex scenarios. &
		Challenges remain in 3D spatial task handling and real-time integration. \\
		
		Liang et al. \cite{170} &
		2023 &
		Introduced ``Code as Policies'' using LLMs to write robot policy code from commands. &
		Enables reactive and trajectory-based control via few-shot prompting. &
		Limited physical-world validation. \\
		
		Xu et al. \cite{171} &
		2025 &
		Reviewed LLM-integrated CPSS frameworks for human-centered Industry 5.0. &
		Connects LLMs, perception, scheduling, and swarm intelligence into unified framework. &
		Lacks empirical task-specific evaluations. \\
		
		Zhao et al. \cite{172} &
		2024 &
		Developed MultiBotGPT system using GPT-3.5 for UAV/UGV task execution. &
		Outperforms BERT in assignment success rate; improves operator experience. &
		Focused on limited command types and single-modal input. \\
		
		Shen et al. \cite{173} &
		2024 &
		Discussed LLMs as foundation for embodied agents in physical environments. &
		Conceptual bridge from virtual to real-world through embodiment. &
		Lacks concrete framework or task-based implementation details. \\
		
		Jiang et al. \cite{174} &
		2024 &
		Proposed KoMA: LLM-driven multi-agent system for autonomous driving. &
		Achieves robust decision-making via planning, shared memory, and reflection. &
		Complexity of real-world deployment and scalability remains a challenge. \\
		
		\bottomrule
	\end{tabularx}
	
\end{table*}

most current approaches rely heavily on pretrained LLMs such as GPT-3.5 or GPT-4, which are optimized for general language tasks but not specifically designed for low-latency, real-time robotic control. This often leads to issues in responsiveness and execution stability, especially in tasks involving tight control loops. Secondly, the majority of frameworks focus on single-agent scenarios or simplified settings, lacking generalizability and scalability to complex, multi-agent, or open-world environments. Thirdly, while some studies explore multi-modal grounding, such as combining visual or force feedback with language, the integration remains shallow—true sensorimotor fusion and real-time adaptive policy updating are still underexplored. In addition, the lack of robust safety constraints and fail-safe mechanisms within LLM-generated policies poses significant risks in high-stakes applications like autonomous driving or human-robot collaboration. Lastly, most systems depend on extensive prompt engineering or human intervention for instruction following, limiting their autonomy and deployability in real-world industrial settings.\par

\section{Discussion and Outlook}
As artificial intelligence continues to evolve, the convergence of large language models (LLMs), structured knowledge bases (KBs), and advanced reasoning capabilities (RA) is shaping a new generation of intelligent agents. While LLMs have demonstrated remarkable performance in language understanding and generation, they are inherently limited in physical interaction, real-time perception, and grounded reasoning. Bridging this gap requires the integration of embodied intelligence (EI), where agents possess not only cognitive abilities but also the capacity to sense, act, and adapt in real-world environments.\par
Moving forward, the development of general embodied intelligence hinges on several critical directions:\par
(1) Lightweight LLMs and Edge Deployment One of the major obstacles to deploying large language models in embodied agents lies in their excessive computational demands. Mainstream LLMs such as GPT-4 often require massive GPU clusters, rendering them impractical for real-time use in robots or embedded systems. Future research must focus on developing lightweight and optimized versions of LLMs, such as distilled, quantized, or sparsely activated models, that maintain core capabilities while significantly reducing memory footprint and inference latency. Additionally, hardware-software co-optimization—including edge AI chips, efficient transformer variants, and low-rank adaptation (LoRA)—will be essential for integrating LLMs into real-time, mobile, and power-constrained environments like home service robots or wearable devices.\par
(2) Closed-loop Knowledge Systems To move beyond static, offline training corpora, LLMs must interface with dynamic and structured knowledge systems. These systems should support real-time retrieval-augmented generation (RAG), knowledge graph reasoning, and contextual updating based on ongoing interactions. Closed-loop knowledge architectures enable agents not only to query factual information but also to update their internal representations based on new evidence or user feedback. Moreover, task-specific memory modules and episodic knowledge buffers can provide grounding for continual learning, allowing embodied agents to recall context, adapt to users, and improve performance over time.\par
(3) Hybrid Reasoning Frameworks General embodied intelligence requires a reasoning mechanism that balances data-driven flexibility with rule-based precision. Hybrid reasoning frameworks offer a promising solution by combining the pattern recognition abilities of LLMs with symbolic reasoning, multi-hop inference, and task planning. For instance, Chain-of-Thought (CoT) reasoning allows models to generate step-by-step intermediate reasoning traces, while Tree-of-Thought (ToT) expands this to explore multiple reasoning branches in parallel. The ReAct framework further integrates reasoning and action, allowing LLMs to interact with external tools or environments while maintaining internal logical consistency. These frameworks are critical for solving complex tasks in uncertain or dynamic physical environments where simple end-to-end prediction fails.\par
(4) Perception-Action Grounding Embodied agents must translate abstract language inputs into concrete physical actions, such as grasping an object or navigating a room. Achieving this requires semantic grounding—the alignment between high-level linguistic\par

commands and low-level sensorimotor capabilities. This can be approached through multi-modal representation learning, policy networks trained via reinforcement learning, and instruction-conditioned behavior cloning. For example, an agent interpreting the instruction “bring me the red cup on the table” must detect objects, resolve spatial references, plan a path, and execute motor commands. Seamless perception-action mapping bridges the gap between symbolic cognition and real-world embodiment.\par
(5) Continual and Interactive Learning Unlike disembodied models trained once and frozen, embodied agents operate in ever-changing environments. Therefore, they must possess the ability for continual learning, where new tasks, scenarios, or users are encountered and learned incrementally without catastrophic forgetting. Moreover, agents must engage in interactive learning, adapting in real time based on dialogue, feedback, or demonstration. This includes techniques such as reinforcement learning with human feedback (RLHF), few-shot adaptation, and lifelong learning architectures. By incorporating online learning loops and personalized adaptation, agents can improve with use, evolve with time, and align more closely with human goals and environments. Ultimately, achieving truly general embodied intelligence will require breaking down the silos between language, knowledge, reasoning, and embodiment. The integration of these components into a unified, adaptive framework holds the key to building agents that can reason, learn, and act autonomously across open, dynamic, and complex real-world settings—marking a pivotal step toward the future of human-level AI.\par

\section{Conclusion}
This paper provides a comprehensive review of the recent advancements in large language models (LLMs), knowledge bases (KBs), reasoning capabilities (RA), and embodied intelligence (EI), highlighting their individual roles and synergistic potential in advancing artificial intelligence. As LLMs evolve from pure text generation engines to multi-modal cognitive agents, their integration with structured knowledge, explicit reasoning mechanisms, and physical embodiment is becoming increasingly crucial. Looking ahead, building a new generation of intelligent agents requires moving beyond isolated modules towards deeply coupled architectures. By aligning LLMs’ powerful language understanding with real-time knowledge access, multi-step reasoning, and interactive embodiment, we can move closer to achieving general embodied intelligence—agents that can perceive, think, learn, and act within open and dynamic environments. The road to this vision is both challenging and promising. It calls for innovations in model compression, knowledge system design, hybrid reasoning frameworks, semantic-action alignment, and lifelong adaptive learning. Through the collaborative development of these key components, we lay the foundation for building next-generation AI agents that are not only linguistically competent but also contextually aware, knowledge-grounded, logically coherent, and physically capable—marking a critical leap toward truly general-purpose artificial intelligence.\par

\section*{Declarations}
All authors declare that they have no conflicts of interest.

\section*{Acknowledgement}
This paper was supported by Innovative Research Group of Chongqing Municipal Education Commission (CXQT19026), and Cooperative Project between Chinese Academy of Sciences and University in Chongqing (HZ2021011). Moreover, this work was supported by the Research Startup Fund of Chongqing University of Technology (0119240197).
Furthermore, this work was supported by equipment funded through the “Intelligent Connected New Energy Vehicle Teaching System” project of Chongqing University of Technology, under the national initiative “Promote large-scale equipment renewals and trade-ins of consumer goods.”
In addition, this document is the results of the research project funded by the Open Foundation of the State Key Laboratory of Fluid Power and Mechatronic Systems under Grant GZKF-202307, and by the Independent Research Project of the State Key Laboratory of Fluid Power and Mechatronic Systems under Grant SKLoFP\_ZZ\_2513, and by the National Natural Science Foundation of China (No. 52505072).

\section*{Author Contributions}

F.Y.: Writing—Original Draft, Writing—Review and Editing, Visualization. 
X.H.: Conceptualization, Visualization. 
L.W.: Data Curation, Conceptualization. 
J.D.: Conceptualization, Project Administration. 
Z.T.: Writing—Review and Editing. 
Y.W.: Data Curation, Conceptualization. 
S.G.: Conceptualization. 
Y.F.: Conceptualization. 
Y.P.: Project Administration, Writing—Review and Editing, Funding Acquisition. 
Z.M.: Project Administration, Writing—Review and Editing, Funding Acquisition. 
All authors have read and agreed to the published version of the manuscript.

\footnotesize\bibliographystyle{IEEEtran}
\bibliography{ref}

\end{document}